\documentclass[lettersize,journal]{IEEEtran}
\usepackage{amsmath,amsfonts}
\usepackage{algorithmic}
\usepackage{algorithm}
\usepackage{array}
\usepackage[caption=false,font=normalsize,labelfont=sf,textfont=sf]{subfig}
\usepackage{textcomp}
\usepackage{stfloats}
\usepackage{url}
\usepackage{verbatim}
\usepackage{graphicx}
\usepackage{cite}
\usepackage{times}
\usepackage{soul}
\usepackage{url}
\usepackage[hidelinks]{hyperref}
\usepackage[utf8]{inputenc}
\usepackage[small]{caption}
\usepackage{graphicx}
\usepackage{amsmath}
\usepackage{amsthm}
\usepackage{booktabs}
\usepackage{algorithm}
\usepackage{algorithmic}
\usepackage[switch]{lineno}
\usepackage{amsfonts}
\usepackage{multirow, makecell}
\usepackage{stfloats}
\usepackage{colortbl}
\usepackage[table]{xcolor}
\begin{document}

\title{DSFNet: Learning Dual-Domain Spectral Operators for Multi-Modality Spatio-Temporal Forecasting in Urban Transportation Systems}

\author{
Yongchao Li, Yang Li, Zhuoxuan Li, Jun Chen, Chu Zhang, Jinde Cao,~\IEEEmembership{Fellow,~IEEE,} and Leszek Rutkowski,~\IEEEmembership{Life~Fellow,~IEEE}%
\thanks{This work was supported in part by the National Natural Science Foundation of China under Grants 52572340, 62576098 and 62573122; in part by the program “Excellence Initiative—Research University” for the AGH University of Krakow; in part by the ARTIQ project under Grants UMO-2021/01/2/ST6/00004 and ARTIQ/0004/2021; and in part by the Polish Ministry of Science and Higher Education funds assigned to the AGH University of Krakow. (Corresponding author: Chu Zhang.)}
\thanks{Yongchao Li, Yang Li, Jun Chen, and Chu Zhang are with the School of Transportation, Southeast University, Nanjing 211189, China, and also with the Jiangsu Province Collaborative Innovation Center of Modern Urban Traffic Technologies, Nanjing 211189, China. (e-mail: 230248216@seu.edu.cn; l\_yang@seu.edu.cn; chenjun@seu.edu.cn; zhangchu0720@seu.edu.cn).}
\thanks{Yang Li is also with the Department of Data Science, City University of Hong Kong, Hong Kong 999077, China.}
\thanks{Zhuoxuan Li is with the School of Mathematics, Southeast University, Nanjing 211189, China, and also with the Systems Research Institute of the Polish Academy of Sciences, 01-447 Warsaw, Poland. (e-mail: 230229338@seu.edu.cn).}
\thanks{Jinde Cao is with the School of Mathematics, Southeast University, Nanjing 211189, China, with the Department of Mathematics, Luoyang Normal University, Luoyang 471934, China, and also with the Purple Mountain Laboratories, Nanjing 211111, China. (e-mail: jdcao@seu.edu.cn).}% <-this % stops a space
\thanks{Leszek Rutkowski is with the Systems Research Institute of the Polish Academy of Sciences, 01-447 Warsaw, Poland, with AGH University of Krakow, 30-059 Kraków, and with the SAN University, 90-113, Łódź, Poland  (e-mail: leszek.rutkowski@ibspan.waw.pl).}
}

\markboth{Journal of \LaTeX\ Class Files,~Vol.~14, No.~8, August~2021}%
{Shell \MakeLowercase{\textit{et al.}}: A Sample Article Using IEEEtran.cls for IEEE Journals}


\maketitle

\begin{abstract}
Multi-Modality Spatio-Temporal Forecasting (MoSTF) extends traditional spatio-temporal forecasting by incorporating diverse traffic modalities. Despite significant recent strides in spatio-temporal modeling, existing approaches often fail to explicitly model the coupling relationships between different modality variables. Accurate MoSTF is challenging, as it requires modeling (1) temporal dynamic heterogeneity under exogenous influences and (2) heterogeneous spatial dependencies alongside complex cross-variable couplings. To address these challenges, we propose the Dual-Domain Spectral Filtering Network (DSFNet). Our framework employs dual-domain spectral filtering to capture heterogeneous spatial patterns and explicitly model the relationships between variables. Unlike graph-based message passing or dense attention over node-modality pairs, DSFNet factorizes space-modality interactions into feature-domain and spatial-domain spectral operators, enabling scalable modeling of nonlocal dependencies and cross-modality couplings. Furthermore, we introduce an external gating mechanism to adaptively regulate temporal dynamics under external influences. We validate our method through extensive experiments on five representative real-world traffic datasets. Compared with the second-best baselines, DSFNet reduces MAE by 3.21\%--10.16\% across these datasets. The results demonstrate that DSFNet significantly outperforms existing state-of-the-art baselines in accuracy while exhibiting efficiency and robustness.
\end{abstract}


\section{Introduction}
\IEEEPARstart{I}{n} transportation systems, interactions among travelers, vehicles, infrastructures, and service facilities continuously generate spatio-temporal traffic states over urban networks. Accurate spatio-temporal forecasting supports a wide range of intelligent transportation applications, such as route planning and traffic control \cite{YANG2026105413,zhang2022forecasting}. Traditionally, these traffic states are often represented by aggregated measurements, such as total traffic volume, speed, or demand at each location and time interval. With the increasing availability of advanced sensing and transaction records, transportation systems can now be observed from multiple perspectives, including movement directions, vehicle types, travel attributes, transaction categories, and user groups. These perspectives define an additional modality domain beyond the spatial and temporal domains, thereby giving rise to multi-modality traffic data \cite{deng2024multi}. Here, ``modality'' denotes semantically distinct traffic observation channels associated with such operational attributes, rather than sensor modalities such as images or LiDAR. Compared with conventional spatio-temporal data that mainly describe aggregated traffic states over space and time, multi-modality traffic data provide a more comprehensive and detailed characterization of transportation operations. However, the introduction of the modality domain also brings high-dimensional structures and heterogeneous dynamics across space, time, and modalities \cite{deng2024multi}. Accordingly, as transportation management increasingly requires more informative and actionable predictions, forecasting models are expected to predict diverse traffic modalities at high spatio-temporal resolution, rather than only estimating aggregated traffic totals.

\begin{figure}[t]
    \centering
    \includegraphics[width=\linewidth]{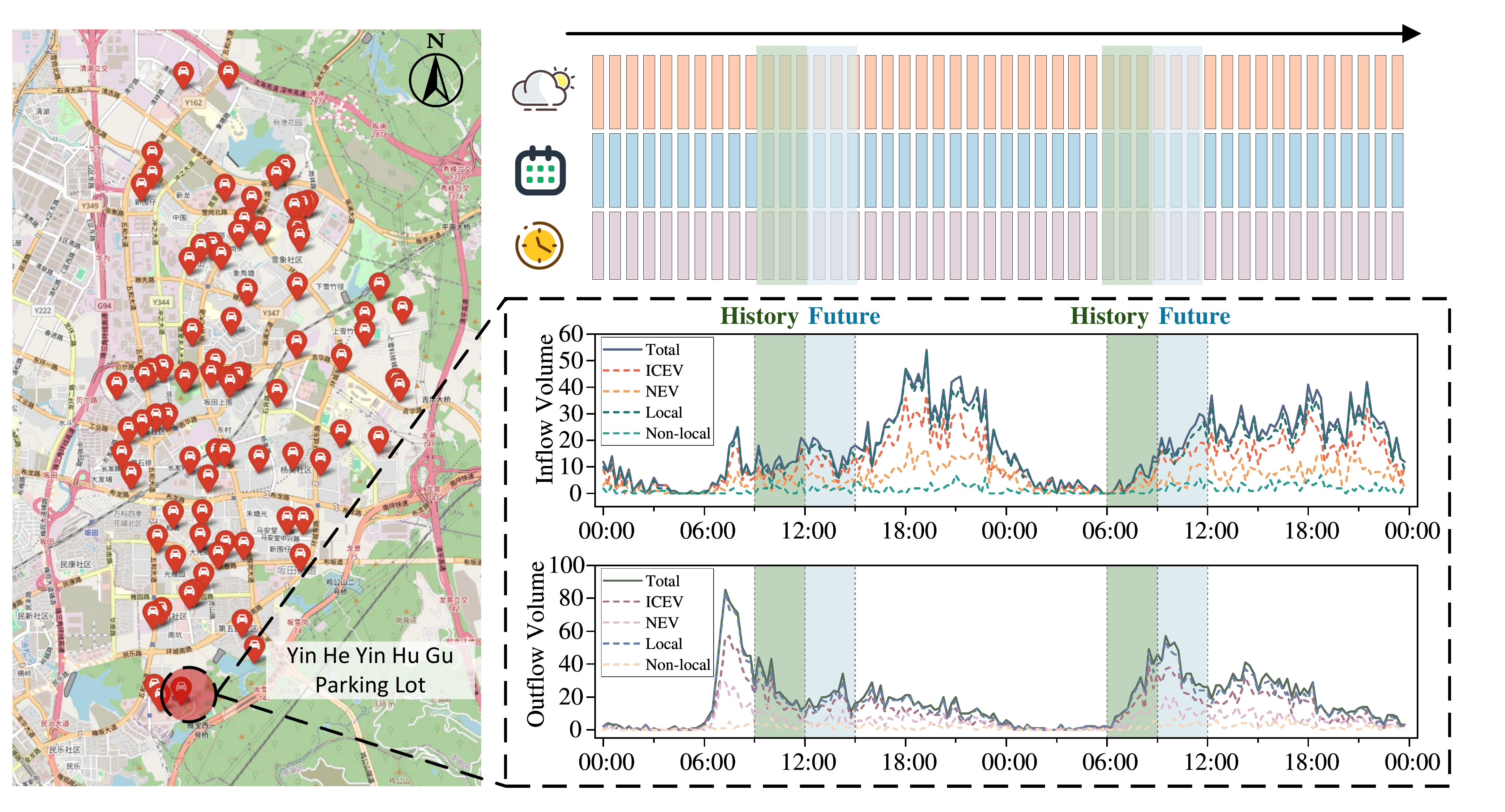}
    \caption{Illustration of a multi-modality urban parking scenario with 83 parking facilities in Shenzhen. The figure illustrates the temporal evolution of different modality-specific parking flows, including inflow/outflow, NEV/ICEV, and local/nonlocal vehicle flows, across multiple facilities. These flows present both shared temporal regularities and modality-dependent heterogeneous dynamics, motivating the need for multi-modality spatio-temporal modeling.}
    \label{fig:figure1}
\end{figure}

To further illustrate this concept, Fig.~\ref{fig:figure1} presents a representative urban parking scenario. In this scenario, traffic volume is represented through multiple meaningful modalities, including inflow and outflow, new energy vehicles (NEV) and internal combustion engine vehicles (ICEV), and local and nonlocal vehicles. These modalities are not independent observations; instead, they exhibit complex correlations and heterogeneous evolution patterns. For instance, inflow and outflow are related through parking duration and travel demand, while NEV and ICEV may respond differently to charging availability and policy incentives. The system dynamics are also influenced by exogenous variables such as weather conditions and holiday schedules \cite{wang2024timexer}. Accurate prediction of multi-modality traffic can assist in facility planning, resource allocation, and operational decision-making \cite{mozaffari2020joint}. Motivated by these demands, this study investigates \emph{Multi-Modality Spatio-Temporal Forecasting} (MoSTF), aiming to jointly model heterogeneous traffic modalities and their spatio-temporal interactions under the influence of exogenous factors.

Achieving MoSTF poses significant modeling challenges. First, target traffic modalities are influenced by various exogenous factors. Holidays may reshape travel demand and mobility patterns, while weather conditions can affect road capacity, parking behavior, and travel preferences. Although methods such as TimeXer \cite{wang2024timexer} integrate exogenous variables via Transformer-based cross-attention to address temporal alignment and irregularity, they usually treat external influences in a shared manner across different traffic modalities. In practice, however, exogenous factors affect traffic modalities heterogeneously, as distinct travel groups, vehicle types, and facilities often exhibit diverse response patterns to the same external disturbance \cite{cui2025weather}. Therefore, further exploration is needed to incorporate external variables into a modality-aware MoSTF framework.

Second, different traffic modalities present complex and nonlinear endogenous interactions rather than evolving as independent parallel series. On highways, passenger-vehicle and freight-vehicle flows interact due to differences in driving behavior and operational constraints \cite{li2025analysis}. In parking systems, inflow and outflow exhibit lagged dependencies shaped by parking duration and travel demand \cite{karaliopoulos2022matching}. Similarly, traffic modalities associated with vehicle energy types, registration attributes, or transaction categories may show correlated yet distinct temporal evolution. Accurately capturing these time-evolving cross-modality relationships is essential for improving forecasting performance.

Finally, traffic modalities exhibit heterogeneous spatial patterns, leading to non-uniform spatial dependencies \cite{ijcai2023p491}. For example, passenger and freight flows may propagate differently due to distinct routing preferences \cite{NADI2024104413}. In urban parking systems, local and nonlocal vehicles may also follow different spatial distributions because of differences in commuting purposes, activity locations, and parking preferences. However, despite this inherent heterogeneity, most graph neural networks (GNNs) adopt a single shared adjacency matrix for all traffic modalities \cite{wu2019graph,han2024adaptive}, implicitly assuming homogeneous spatial relations. Existing methods still lack an explicit mechanism to model modality-specific spatial dependencies in multi-modality transportation systems.

In this study, we propose the \textbf{D}ual-Domain \textbf{S}pectral \textbf{F}iltering \textbf{Net}work \textbf{(DSFNet)} for MoSTF. At its core, DSFNet introduces a \emph{Dual-Domain Interaction Block} (DDIBlock), built upon the \emph{Graph Cosine Operator} (GCO) \cite{zhang2024predicting}. The DDIBlock performs spectral filtering in both the feature domain and the spatial domain. It explicitly models cross-modality correlations and reconstructs heterogeneous spatial dependency patterns among traffic modalities. DSFNet employs causal dilated convolutions to extract temporal dependencies efficiently \cite{yu2015multi}. In addition, an exogenous-variable-driven gating mechanism is introduced to adaptively regulate information flow under varying external conditions \cite{lan2025gateformer}. Together, these designs form a unified and scalable framework that jointly addresses modality-specific spatial dynamics, cross-modality interactions, and exogenous-aware temporal forecasting.

In summary, our contributions are threefold:

(1) We propose a unified framework, DSFNet, for MoSTF. Its core module, DDIBlock, performs dual-domain spectral filtering to explicitly model cross-modality interactions and heterogeneous spatial dependencies. Unlike shared-graph approaches, this mechanism learns modality-specific spatial patterns and reconstructs cross-modality couplings, effectively alleviating the limitations of homogeneous dependency modeling.

(2) We devise an Exogenous-Gated Dilated Convolution Block (EGDCBlock) for adaptive temporal modeling. By integrating external factors into a learnable gating mechanism, this module dynamically modulates temporal evolution across different traffic modalities, enhancing robustness under complex environmental shifts.

(3) We conduct extensive experiments on five representative real-world transportation datasets. The results demonstrate that DSFNet consistently outperforms state-of-the-art baselines in accuracy, while exhibiting superior computational efficiency and robustness against noise and missing data.

The rest of this paper is organized as follows. Section~\ref{sec:related_work} reviews related studies on spatio-temporal traffic forecasting, Transformer-based forecasting, and spectral spatio-temporal modeling. Section~\ref{sec:Methodology} formulates the MoSTF problem and presents the proposed DSFNet framework. Section~\ref{sec:experiments} reports extensive experiments on five real-world transportation datasets, together with ablation, efficiency, and robustness analyses. Finally, Section~\ref{sec:Conclusion and Future Work} concludes this paper and discusses future research directions.

\section{Related Work}
\label{sec:related_work}
In this section, we provide a brief review of existing studies related to DSFNet. Specifically, we first introduce STGNN-based spatio-temporal traffic forecasting methods. We then review Transformer-based forecasting models and their applications to multi-modality spatio-temporal forecasting. Finally, we discuss Fourier neural operators and spectral modeling approaches, highlighting their distinction from graph-based message passing and their relevance to dual-domain interaction modeling.

\subsection{STGNN-based Spatio-Temporal Forecasting}
Spatio-temporal traffic forecasting is a fundamental task in intelligent transportation systems, aiming to predict future traffic states from historical observations over transportation networks. Unlike generic time-series forecasting, it requires modeling both temporal dynamics within each traffic series and spatial dependencies among transportation entities, such as sensors, road segments, regions, and service facilities. Since such dependencies are often irregular and non-Euclidean, traffic forecasting requires models that can jointly capture temporal evolution and network-level spatial correlation.

Early deep learning methods commonly employed CNNs to extract spatial patterns and combined them with RNNs or TCNs for temporal modeling. However, grid-based CNNs are limited in representing irregular transportation networks. With the development of graph convolutional networks (GCNs), spatio-temporal graph neural networks (STGNNs) have become a dominant paradigm. STGNNs typically encode spatial dependencies using predefined graphs derived from road connectivity, geographical distance, or functional similarity, and integrate graph convolution with temporal modules. Representative examples include DCRNN \cite{li2018dcrnn_traffic}, which combines diffusion graph convolution with recurrent units; STGCN \cite{yu2018spatio}, which integrates graph convolution and gated temporal convolution; GWNet \cite{wu2019graph}, which introduces adaptive graph learning with dilated temporal convolutions; and MTGNN \cite{wu2020connecting}, which learns latent graph structures for multivariate forecasting. Attention mechanisms and automated graph design have also been incorporated into STGNNs, as in GMAN \cite{zheng2020gman}, ASTGNN \cite{guo2021learning}, and AutoSTF \cite{lyu2025autostf}. More recent studies, such as STPGNN \cite{kong2024spatio} and AGCNDE \cite{han2024adaptive}, further investigate pivotal graph structures and continuous spatio-temporal dynamics.

Despite their effectiveness, STGNNs still face several limitations. Predefined graphs may be biased or unavailable, as physical connectivity or geographical proximity cannot fully reflect complex traffic interactions. Latent graph learning methods, such as AGCRN \cite{bai2020adaptive}, MTGNN \cite{wu2020connecting}, GTS \cite{shang2021discrete}, StemGNN \cite{cao2020spectral}, and DFDGCN \cite{li2024dynamic}, alleviate this issue by inferring graph structures from data. However, both predefined-graph and latent-graph STGNNs generally rely on graph convolution or message passing, whose cost increases with the number of spatial entities and graph edges. Efficient linear or hybrid models, such as DLinear \cite{zeng2023transformers} and TiDE \cite{das2023longterm}, reduce computational overhead but often weaken explicit spatial topology and cross-variable interaction modeling. More importantly, most existing methods adopt a shared graph structure, unified propagation mechanism, or common latent representation for all traffic variables. This assumption is restrictive for MoSTF, where different traffic modalities may exhibit heterogeneous spatial propagation patterns and modality-specific interactions.

\subsection{Transformer-based Spatio-Temporal Forecasting}

Transformer-based models have been widely explored for time-series and spatio-temporal forecasting due to their flexible attention mechanisms. Compared with recurrent models based on sequential state transitions or convolutional models with fixed receptive fields, Transformers directly establish dependencies across time steps, variables, and contextual factors. Representative forecasting models include Informer \cite{zhou2021informer}, which improves attention efficiency; Autoformer \cite{wu2021autoformer}, which introduces series decomposition and auto-correlation; FEDformer \cite{zhou2022fedformer}, which incorporates frequency-domain representations; Crossformer \cite{zhang2023crossformer}, which captures cross-dimension dependencies; iTransformer \cite{liu2024itransformer}, which treats variables as tokens; and TimeXer \cite{wang2024timexer}, which incorporates exogenous variables through cross-attention. These methods demonstrate the effectiveness of attention mechanisms in modeling temporal dependencies, cross-variable relationships, and external influences.

Recent studies have further extended Transformer-based modeling to traffic forecasting and MoSTF. MoSTF introduces an additional modality domain beyond the spatial and temporal domains, requiring models to jointly capture temporal evolution, spatial dependencies, and modality interactions \cite{deng2024multi}. In this context, attention mechanisms are attractive because they can flexibly model interactions among heterogeneous variables and contextual factors. For example, STHGFormer \cite{li2024towards} studies integrated traffic forecasting with heterogeneous graph-enhanced Transformer structures, while MoSSL \cite{deng2024multi} employs modality-aware attention and self-supervised learning to capture latent dependencies from temporal, spatial, and modality perspectives. Related studies on detailed traffic inference \cite{li2025fine} and multi-view traffic modeling also highlight the need to exploit diverse traffic observations beyond aggregate volume prediction. Exogenous-aware models, such as TimeXer \cite{wang2024timexer}, further provide useful mechanisms for incorporating external variables through cross-attention.

Nevertheless, Transformer-based models remain limited for scalable MoSTF. Generic time-series Transformers usually lack explicit spatial topology inductive bias, which is essential for modeling traffic propagation over irregular transportation networks. Shared attention layers and embeddings may also entangle heterogeneous modalities and weaken modality-specific representations. Moreover, treating each node-modality pair as a token can lead to quadratic complexity $O((NC)^2)$, where $N$ and $C$ denote the numbers of spatial entities and traffic modalities, respectively. Existing exogenous-aware Transformers also tend to model external factors in a shared manner, overlooking their heterogeneous effects across modalities and locations.

\subsection{Fourier Neural Operators and Spectral Spatio-Temporal Modeling}

Fourier neural operators (FNOs) provide an efficient spectral framework for learning mappings between function spaces. By parameterizing integral kernels in the Fourier domain, FNOs capture global dependencies through fast spectral transformations, making them suitable for modeling long-range spatio-temporal dynamics \cite{li2020fourier}. Owing to these advantages, FNOs and their variants have been applied to various scientific and engineering tasks, including turbulent flow simulation, geophysical forecasting, weather prediction, and dynamic wake modeling, as demonstrated by FNO-DST \cite{nag2026spatio}, ST-FNO \cite{zhang2025novel}, and FANO \cite{lin2026fano}.

Spectral modeling has also been increasingly explored in time-series forecasting. Time signals usually contain periodic demand fluctuations, trend components, high-frequency disturbances, and long-range spatial correlations, which can be effectively characterized in the spectral domain. This provides a modeling perspective different from both attention-based and graph-convolution-based approaches. While attention mechanisms usually learn dependencies by computing pairwise similarities among tokens in the data domain, spectral operators transform signals into a global basis space and model dependencies by modulating informative frequency components. Similarly, instead of propagating information locally along predefined or learned edges as in graph convolution, Fourier-style operators can capture nonlocal interactions through global spectral modes. Therefore, spectral modeling offers an efficient way to represent periodicity, global dependency, and nonlocal correlation, which are common in large-scale transportation systems.

This perspective has motivated a series of spectral forecasting models. For general time-series forecasting, FEDformer \cite{zhou2022fedformer} introduces frequency-enhanced representations into Transformer-based forecasting, FilterNet \cite{yi2024filternet} employs learnable frequency filters to selectively preserve informative components, and TSLANet \cite{eldele2024tslanet} replaces self-attention with adaptive spectral modeling to capture both long- and short-range dependencies. For traffic-related forecasting, FourierGNN \cite{yi2023fouriergnn} performs graph operations in Fourier space to jointly model spatial and temporal dependencies with reduced complexity. DeepPA further adopts a graph cosine operator based on discrete cosine transform to capture global spatial dependencies among large-scale parking lots \cite{zhang2024predicting}. These studies show that spectral operators can serve as an efficient alternative to dense attention and local message passing, especially when traffic dependencies are global, nonlocal, or not fully determined by physical distance and road connectivity.

However, existing spectral forecasting methods mainly focus on temporal-frequency modeling or spatial dependency modeling, while rarely considering the joint structure of spatial entities and traffic modalities. Directly applying full spectral or attention operators to all node-modality pairs is computationally expensive. Therefore, an efficient structured spectral operator is needed to separately but jointly model modality-wise and spatial-wise dependencies in MoSTF.

\section{Methodology}
\label{sec:Methodology}
In this section, we introduce the proposed DSFNet framework for MoSTF. We first define the MoSTF problem and then present the overall architecture. After that, we detail two core modules: the EGDCBlock, which models exogenous-aware temporal dynamics, and the DDIBlock, which captures cross-modality interactions and heterogeneous spatial dependencies via dual-domain spectral filtering.

\subsection{Problem Settings}
The transportation system is defined to consist of $N$ spatial entities. At each time step $t$, each entity observes a vector of traffic variables, forming a multi-modality traffic state 
$\mathbf{X}^{t}=\{\mathbf{x}^{t}_{1},\mathbf{x}^{t}_{2},\ldots,\mathbf{x}^{t}_{N}\}\in\mathbb{R}^{N\times F}$, 
where $F$ denotes the number of traffic modalities, including the total volume and modality-specific traffic observations. 
$\mathbf{x}^{t}_{n}\in\mathbb{R}^{F}$ represents the traffic state vector of entity $n$ at time $t$. 
Correspondingly, each entity is associated with an exogenous variable vector 
$\mathbf{Z}^{t}=\{\mathbf{z}^{t}_{1},\mathbf{z}^{t}_{2},\ldots,\mathbf{z}^{t}_{N}\}\in\mathbb{R}^{N\times F_{\mathrm{ex}}}$, 
where $F_{\mathrm{ex}}$ is the dimensionality of exogenous variables and 
$\mathbf{z}^{t}_{n}\in\mathbb{R}^{F_{\mathrm{ex}}}$ denotes the exogenous feature vector of entity $n$ at time $t$. 

Given the historical multi-modality traffic states over the past $T$ time steps, we denote the input as 
$\mathbf{X}_{1:T}\in\mathbb{R}^{T\times N\times F}$, and the corresponding exogenous variables as 
$\mathbf{Z}_{1:T}\in\mathbb{R}^{T\times N\times F_{\mathrm{ex}}}$. 
The objective is to learn a predictive function $\mathcal{F}_{\theta}(\cdot)$, parameterized by $\theta$, that estimates future traffic states over the next $S$ steps:
\begin{equation}
    \widehat{\mathbf{X}}_{T+1:T+S} = \mathcal{F}_{\theta}(\mathbf{X}_{1:T}, \mathbf{Z}_{1:T}).
\end{equation}
The forecasting model needs to simultaneously capture 
(i) nonlinear endogenous interactions among traffic modalities, 
(ii) heterogeneous spatial dependencies associated with different modalities, and 
(iii) temporal variations driven by exogenous variables.

\begin{figure*}[t]
    \centering
    \includegraphics[width=\textwidth]{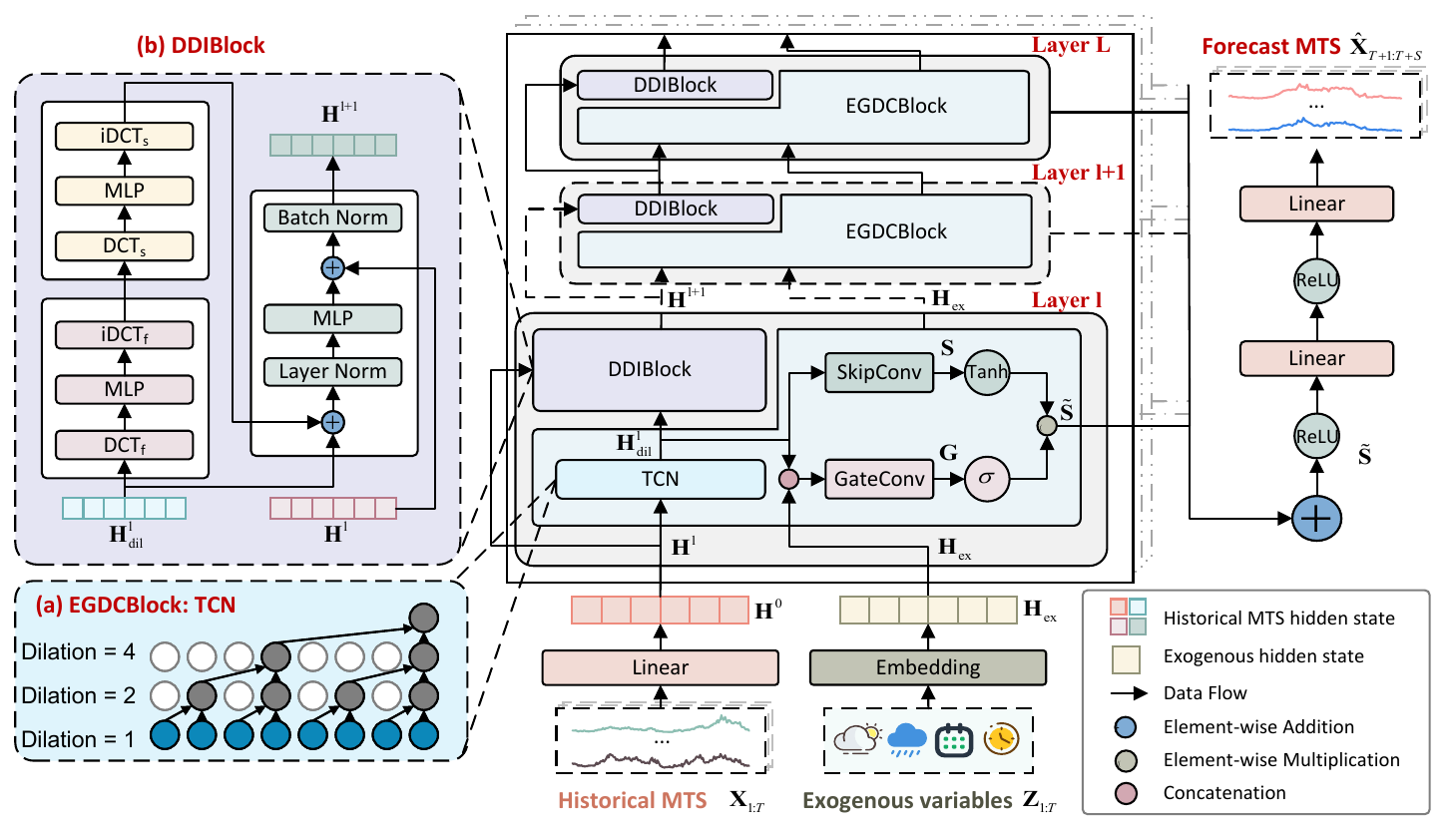}
    \caption{Overall framework of the proposed DSFNet. Historical traffic series and exogenous variables are encoded into latent representations. Each layer contains an Exogenous-Gated Dilated Convolution Block (EGDCBlock) for exogenous-aware temporal modeling with gated skip connections, followed by a Dual-Domain Interaction Block (DDIBlock) for spectral interaction modeling in spatial and feature domains. Skip outputs from all layers are aggregated to produce MoSTF.}
    \label{fig:framework}
\end{figure*}

\subsection{Overall Framework}
We develop DSFNet as illustrated in Figure~\ref{fig:framework}. The model takes as input the historical series $\mathbf{X}_{1:T}$ and the corresponding exogenous variables $\mathbf{Z}_{1:T}$. The historical multi-modality traffic series $X_{1:T}$ is first projected 
from the original modality space $\mathbb{R}^{F}$ into a latent 
modality-aware channel space $\mathbb{R}^{C}$ through a linear embedding 
layer. In this way, each hidden channel is learned from the original traffic 
modalities and can encode modality-related patterns. The exogenous variables 
$Z_{1:T}$ are encoded using learnable embedding layers. The embedded 
traffic and exogenous representations are then processed by $L$ stacked 
DSFNet layers.

\paragraph{Exogenous-Gated Dilated Convolution Block (EGDCBlock)}
The EGDCBlock serves as the temporal modeling component. It first applies causal dilated convolutions to capture multi-scale temporal dependencies. Instead of standard residual updates, the block adopts a parameterized skip-connection mechanism that allows stable gradient propagation across layers \cite{wu2019graph}. To incorporate external influences, the exogenous embeddings are concatenated with the convolutional outputs and passed through learnable gating units. These gates dynamically modulate the contribution of skip pathways, enabling the model to adjust temporal information flow under varying external conditions such as weather shifts or holiday effects.

\paragraph{Dual-Domain Interaction Block (DDIBlock)}
After temporal processing, the hidden representations are passed to the DDIBlock, which models spatial dependencies and cross-modality interactions through dual-domain spectral filtering. 
Specifically, feature-domain filtering is used to capture interactions among traffic modalities, while spatial-domain filtering is used to capture global dependencies among transportation entities. 
Rather than explicitly constructing a dense joint operator over all node-modality pairs, DDIBlock adopts a structured dual-domain approximation, which improves scalability while retaining the ability to model both modality interactions and heterogeneous spatial patterns.

\subsection{Exogenous-Gated Dilated Convolution Block (EGDCBlock)}
EGDCBlock expands the temporal receptive field exponentially without increasing model depth and further integrates an exogenous-aware gating mechanism to adapt temporal contributions to varying external conditions.

Given the hidden representation $\mathbf{H}^l \in \mathbb{R}^{C \times N \times T}$ at layer $l$, EGDCBlock first applies a dilated causal convolution to extract multi-scale temporal patterns. By skipping input positions with a dilation factor $d$, the model captures long-range temporal dependencies in a non-recursive fashion:
\begin{equation}
    \mathbf{H}^{l}_{dil} = \mathrm{Conv}_{\mathrm{dil}}\!\left(\mathbf{H}^{l}\right).
\end{equation}
where the dilation factor grows exponentially across layers (e.g., $1, 2, 4, \ldots$). This design enables efficient temporal modeling and alleviates the gradient instability issues commonly associated with recurrent architectures.

Following this, we introduce a parameterized skip connection explicitly conditioned on exogenous features. This allows the model to dynamically adjust the contribution of historical information according to external contexts such as weather or time-of-day patterns.

The encoded exogenous variables are represented as hidden states 
$\mathbf{H}_{\mathrm{ex}} \in \mathbb{R}^{C_{\mathrm{ex}} \times N \times T}$.
We first obtain a skip candidate via a $1{\times}1$ convolution:
\begin{equation}
    \mathbf{S} = \mathrm{Conv}_{skip}(\mathbf{H}^{l}_{dil}).
\end{equation}
Then, we concatenate $\mathbf{H}^{l}_{dil}$ with $\mathbf{H}_{\mathrm{ex}}$ to compute a learnable gating signal:
\begin{equation}
    \mathbf{G} = \sigma\!\left( \mathrm{Conv}_{gate}([\mathbf{H}^{l}_{dil}, \mathbf{H}_{\mathrm{ex}}]) \right),
\end{equation}
where $\sigma(\cdot)$ denotes the sigmoid activation.
The gated skip output is expressed as:
\begin{equation}
    \widetilde{\mathbf{S}} = \operatorname{Tanh}(\mathbf{S}) \odot \mathbf{G}.
\end{equation}
Finally, the skip output of this layer is accumulated with that of the previous layer:
\begin{equation}
    \mathbf{S}^{l}
        = \mathbf{S}^{l-1} + \widetilde{\mathbf{S}}.
\end{equation}

\subsection{Dual-Domain Interaction Block (DDIBlock)}
The DDIBlock is designed to capture interactions that are difficult to represent with a shared graph structure or a shared latent representation. 
Inspired by GCO \cite{zhang2024predicting}, we adopt a Laplacian-inspired contrastive interaction formulation as a structured inductive prior to describe pairwise variations in hidden states, rather than imposing a strict physical heat-diffusion law on traffic systems. 
Different from GCO, which focuses on spatial interaction modeling, DDIBlock extends spectral filtering to both the spatial and feature domains, enabling a structured dual-domain approximation of spatial dependencies and cross-modality relationships.

\begin{figure*}[t]
    \centering
    \includegraphics[width=\textwidth]{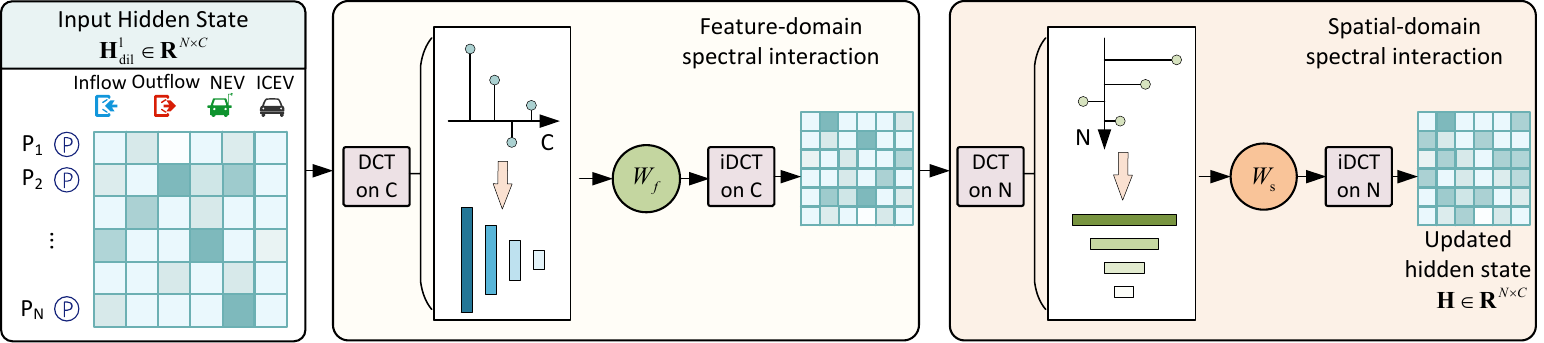}
   \caption{Illustration of the dual-domain spectral operation in DDIBlock. The feature-domain operator $\mathcal{D}_{f}$ first performs DCT-based filtering along the modality dimension to capture cross-modality interactions, and the spatial-domain operator $\mathcal{D}_{s}$ then performs DCT-based filtering along the spatial-entity dimension to model spatial dependencies. The parking-lot case is used as a representative example. The figure focuses on the interaction term $\mathcal{D}_{s}(\mathcal{D}_{f}(\mathbf{H}^{l}_{dil}))$; the residual connection is applied later in the layer-wise update.}
    \label{fig:ddi}
\end{figure*}

As illustrated in Fig.~\ref{fig:ddi}, DDIBlock adopts a serial dual-domain structure. 
The hidden representation is first processed by the feature-domain spectral operator to model cross-modality interactions, and the resulting intermediate representation is then fed into the spatial-domain spectral operator to capture dependencies across parking lots. 
Finally, the dual-domain output is combined with the original hidden representation through a residual connection.

\paragraph{Spatial Operator (Node-wise Interaction)}
We first consider spatial interactions among transportation entities. 
At each time step $t$, let $\hat{\mathbf{H}}_{s}^{t}\in\mathbb{R}^{N\times C}$ denote the hidden representation of all spatial entities. 
We describe how spatial dependencies induce variations in hidden states. 
It should be noted that this formulation is used as an inductive modeling form rather than a strict physical heat-diffusion law. 
Specifically, the variation of entity $i$ is represented as a weighted aggregation of contrastive differences from other entities:
\begin{equation}
    \Delta \hat{\mathbf{H}}^{t}_{s,i}
    =
    \sum_{j=1}^{N} k_{ji}
    \left(
    \hat{\mathbf{H}}^{t}_{s,j}
    -
    \hat{\mathbf{H}}^{t}_{s,i}
    \right),
\end{equation}
where $k_{ji}$ denotes a learnable spatial interaction coefficient from entity $j$ to entity $i$. 
Different from a physical thermodynamic coefficient, $k_{ji}$ is learned from traffic data and measures the dependency strength between transportation entities.

Stacking the variations of all entities gives:
\begin{equation}
    \Delta \hat{\mathbf{H}}^{t}_{s}
    =
    \hat{\mathbf{K}}_{s}\hat{\mathbf{H}}^{t}_{s}
    -
    \hat{\mathbf{D}}_{s}\hat{\mathbf{H}}^{t}_{s},
\end{equation}
where $\hat{\mathbf{K}}_{s}\in\mathbb{R}^{N\times N}$ collects the estimated spatial interaction coefficients, and 
$\hat{\mathbf{D}}_{s}$ is the corresponding estimated diagonal degree matrix with 
$(\hat{\mathbf{D}}_{s})_{ii}=\sum_{j}k_{ji}$. 
The two matrices play roles analogous to an adjacency matrix and a degree matrix, respectively. 
Thus, the above variation can be written in a Laplacian-like form:
\begin{equation}
    \Delta \hat{\mathbf{H}}^{t}_{s}
    =
    (\hat{\mathbf{K}}_{s}-\hat{\mathbf{D}}_{s})\hat{\mathbf{H}}^{t}_{s}
    =
    \hat{\mathbf{L}}_{s}\hat{\mathbf{H}}^{t}_{s},
\end{equation}
where $\hat{\mathbf{L}}_{s}=\hat{\mathbf{K}}_{s}-\hat{\mathbf{D}}_{s}$ denotes the estimated spatial interaction operator.

We further relate the estimated spatial interaction operator to spectral filtering for efficient global dependency modeling. 
Let $\mathbf{L}_{s}$ represent the reference Laplacian matrix, and we aim to fit the estimated operator $\hat{\mathbf{L}}_{s}$ by adding a set of trainable weights $\sigma_{s}$ to $\mathbf{L}_{s}$. 
Thus, the spatial interaction can be written as:
\begin{equation}
    \hat{\mathbf{L}}_{s}\hat{\mathbf{H}}^{t}_{s}
    =
    \sigma_{s}\mathbf{L}_{s}\hat{\mathbf{H}}^{t}_{s}.
\end{equation}

Let $\lambda_{s}$ and $\mathbf{v}_{s}$ be the eigenvalues and eigenvectors of $\mathbf{L}_{s}$, respectively. 
Then the above operation can be expressed as:
\begin{equation}
    \hat{\mathbf{L}}_{s}\hat{\mathbf{H}}^{t}_{s}
    =
    \sigma_{s}\mathbf{L}_{s}\mathbf{v}_{s}
    =
    \sigma_{s}\lambda_{s}\mathbf{v}_{s}.
\end{equation}

Left-multiplying by the Laplacian matrix is equivalent to solving for the second derivative \cite{koren2003spectral}. 
Therefore, for Fourier bases, the transformation can be denoted as:
\begin{equation}
    \sigma_{s}\mathbf{L}_{s}e^{-j\omega t}
    =
    \sigma_{s}
    \frac{\partial^{2}e^{-j\omega t}}{\partial t^{2}}
    =
    \sigma_{s}(-\omega^{2})e^{-j\omega t},
\end{equation}
where $-\omega^{2}$ represents the eigenvalue of the Laplacian matrix in the Fourier basis.

Defining the set of eigenvalues of $\mathbf{L}_{s}$ as:
\begin{equation}
    \boldsymbol{\Lambda}_{s}
    =
    (\lambda_{s,1},\lambda_{s,2},\ldots,\lambda_{s,N})
    \in (-\omega^{2}),
\end{equation}
and the corresponding unit eigenvectors as:
\begin{equation}
    \mathbf{U}_{s}
    =
    (\vec{\mathbf{u}}_{s,1},\vec{\mathbf{u}}_{s,2},\ldots,\vec{\mathbf{u}}_{s,N})
    \in e^{-j\omega t},
\end{equation}
the spatial interaction operator can be rewritten as:
\begin{equation}
    \hat{\mathbf{L}}_{s}\hat{\mathbf{H}}^{t}_{s}
    =
    \sigma_{s}
    \mathbf{U}_{s}\boldsymbol{\Lambda}_{s}\mathbf{U}_{s}^{\top}
    \hat{\mathbf{H}}^{t}_{s}
    =
    \sigma_{s}
    \mathcal{F}^{-1}
    \left(
        \mathcal{F}(\boldsymbol{\Lambda}_{s})
        \odot
        \mathcal{F}(\hat{\mathbf{H}}^{t}_{s})
    \right),
\end{equation}
where $\mathcal{F}(\cdot)$ denotes the Discrete Fourier Transform (DFT). 
Inspired by Fourier Neural Operator (FNO) \cite{li2020fourier}, the trainable weights $\sigma_{s}$ and the spectral response $\mathcal{F}(\boldsymbol{\Lambda}_{s})$ can be parameterized by neural layers, enabling data-driven approximation of global spatial interactions.

However, DFT involves complex-valued computation. 
Following the efficiency consideration in GCO \cite{zhang2024predicting}, we implement the spectral operator using the Discrete Cosine Transform (DCT), which operates on real-valued coefficients and reduces computational overhead. 
Accordingly, the spatial spectral operator is implemented as:
\begin{equation}
    \mathcal{D}_{s}(\hat{\mathbf{H}}^{t}_{s})
    =
    \mathrm{MLP}_{s}
    \left(
        \mathcal{C}_{s}^{-1}
        \left(
        \hat{\mathbf{W}}_{s}
        \odot
        \mathcal{C}_{s}(\hat{\mathbf{H}}^{t}_{s})
        \right)
    \right),
\end{equation}
where $\mathcal{C}_{s}$ and $\mathcal{C}_{s}^{-1}$ denote the forward and inverse DCT along the spatial dimension, and 
$\hat{\mathbf{W}}_{s}$ is a learnable spatial spectral filter. 
This DCT-based implementation provides an efficient real-valued spectral approximation for global spatial interaction modeling, without explicitly constructing dense pairwise propagation over all entities.

\paragraph{Feature Operator (Latent Modality-wise Interaction)}

Besides spatial dependencies, MoSTF also requires modeling interactions among traffic modalities. 
For a fixed entity $i$ at time step $t$, let $\hat{\mathbf{H}}^{t}_{f,i}\in\mathbb{R}^{1\times C}$ denote its feature-domain hidden representation, where the feature dimension corresponds to traffic modalities or latent modality-aware channels. 
Analogous to the spatial operator, we use a Laplacian-inspired contrastive 
interaction formulation to describe how each channel varies under the influence of other channels:
\begin{equation}
    \Delta \hat{H}^{t}_{i,c}
    =
    \sum_{k=1}^{C}
    w_{kc}
    \left(
    \hat{H}^{t}_{i,k}
    -
    \hat{H}^{t}_{i,c}
    \right),
\end{equation}
where $w_{kc}$ denotes a learnable interaction coefficient from channel $k$ to channel $c$. 
Here, $w_{kc}$ measures cross-modality dependency strength rather than any physical diffusion coefficient.

For entity $i$, the above formulation can be rewritten in matrix form as:
\begin{equation}
    \Delta \hat{\mathbf{H}}^{t}_{f,i}
    =
    \hat{\mathbf{H}}^{t}_{f,i}\hat{\mathbf{W}}_{f}
    -
    \hat{\mathbf{H}}^{t}_{f,i}\hat{\mathbf{D}}_{f},
\end{equation}
where $\hat{\mathbf{W}}_{f}\in\mathbb{R}^{C\times C}$ collects the estimated cross-channel interaction coefficients, and 
$\hat{\mathbf{D}}_{f}$ is the corresponding estimated diagonal degree matrix with 
$(\hat{\mathbf{D}}_{f})_{cc}=\sum_{k}w_{kc}$. 
Thus, the feature-domain variation can be written as:
\begin{equation}
    \Delta \hat{\mathbf{H}}^{t}_{f,i}
    =
    \hat{\mathbf{H}}^{t}_{f,i}
    (\hat{\mathbf{W}}_{f}-\hat{\mathbf{D}}_{f})
    =
    \hat{\mathbf{H}}^{t}_{f,i}\hat{\mathbf{L}}_{f},
\end{equation}
where $\hat{\mathbf{L}}_{f}=\hat{\mathbf{W}}_{f}-\hat{\mathbf{D}}_{f}$ denotes the estimated feature-domain interaction operator.

Stacking all entities yields:
\begin{equation}
    \Delta \hat{\mathbf{H}}^{t}_{f}
    =
    \hat{\mathbf{H}}^{t}_{f}\hat{\mathbf{L}}_{f}.
\end{equation}
Different from the spatial operator, which acts on the left side of the hidden representation, the feature operator acts by right multiplication, indicating that the interaction is performed along the feature dimension.

Similar to the spatial operator, we approximate the estimated feature-domain operator by applying learnable spectral scaling to a reference feature-domain Laplacian matrix $\mathbf{L}_{f}$:
\begin{equation}
    \hat{\mathbf{H}}^{t}_{f}\hat{\mathbf{L}}_{f}
    =
    \hat{\mathbf{H}}^{t}_{f}\sigma_{f}\mathbf{L}_{f},
\end{equation}
where $\sigma_f$ denotes learnable feature-domain scaling coefficients. 
Let $\boldsymbol{\Lambda}_{f}$ and $\mathbf{U}_{f}$ be the eigenvalues and eigenvectors of $\mathbf{L}_{f}$. 
Then:
\begin{equation}
    \hat{\mathbf{H}}^{t}_{f}\hat{\mathbf{L}}_{f}
    =
    \sigma_{f}
    \hat{\mathbf{H}}^{t}_{f}
    \mathbf{U}_{f}\boldsymbol{\Lambda}_{f}\mathbf{U}_{f}^{\top}.
\end{equation}

From the spectral filtering view, this operation can be expressed as:
\begin{equation}
    \hat{\mathbf{H}}^{t}_{f}\hat{\mathbf{L}}_{f}
    =
    \sigma_{f}
    \mathcal{F}_{f}^{-1}
    \left(
        \mathcal{F}(\hat{\mathbf{H}}^{t}_{f})
        \odot
        \mathcal{F}(\boldsymbol{\Lambda}_{f})
    \right),
\end{equation}
where $\mathcal{F}$ and $\mathcal{F}^{-1}$ denote the forward and inverse spectral transforms along the feature dimension.

In implementation, we instantiate the feature-domain spectral transform using DCT for efficiency. 
Thus, the feature-domain operator is implemented as:
\begin{equation}
    \mathcal{D}_{f}(\hat{\mathbf{H}}^{t}_{f})
    =
    \mathrm{MLP}_{f}
    \left(
        \mathcal{C}_{f}^{-1}
        \left(
            \mathcal{C}_{f}(\hat{\mathbf{H}}^{t}_{f})
            \odot
            \hat{\mathbf{W}}^{\prime}_{f}
        \right)
    \right),
\end{equation}
where $\mathcal{C}_{f}$ and $\mathcal{C}_{f}^{-1}$ denote the forward and inverse DCT along the feature dimension, and 
$\hat{\mathbf{W}}^{\prime}_{f}$ is a learnable feature-domain spectral filter. 
This DCT-based implementation provides an efficient approximation to global cross-modality interaction modeling while avoiding dense cross-channel attention. The feature-domain operator in DSFNet is not directly applied 
to the raw modality dimension $F$, but to the latent channel space derived 
from multi-modality traffic observations. These latent channels are learned 
from and supervised by the original modality-specific targets, and thus 
serve as modality-aware representations for modeling cross-modality 
dependencies.

\paragraph{Dual-Domain Integration}

After defining the spatial-domain and feature-domain operators, we integrate them through a structured dual-domain approximation. 
For a hidden representation $\hat{\mathbf{H}}^{t}\in\mathbb{R}^{N\times C}$, a fully joint operator over all node-modality states would require modeling dependencies in an $NC$-dimensional space, which is computationally expensive for large-scale MoSTF. 
Therefore, instead of explicitly constructing a dense joint operator, we approximate the joint variation by combining spatial and feature interactions:
\begin{equation}
    \Delta \hat{\mathbf{H}}^{t}
    =
    \hat{\mathbf{L}}_{s}
    \hat{\mathbf{H}}^{t}
    \hat{\mathbf{L}}_{f},
\end{equation}
where $\hat{\mathbf{L}}_{s}$ acts on the spatial dimension and $\hat{\mathbf{L}}_{f}$ acts on the feature dimension. 
This formulation provides a separable approximation of the joint space-modality interaction.

In practice, we implement this operation using the DCT-based spatial and feature operators derived above:
\begin{equation}
    \Delta \hat{\mathbf{H}}^{t}
    \approx
    \mathcal{D}_{s}
    \left(
        \mathcal{D}_{f}
        \left(
            \hat{\mathbf{H}}^{t}
        \right)
    \right).
\end{equation}
Here, $\mathcal{D}_{f}$ captures cross-modality interactions along the feature dimension, while $\mathcal{D}_{s}$ captures spatial interactions along the entity dimension. 
Under a residual learning scheme, the layer-wise update is:
\begin{equation}
    \mathbf{H}^{l+1}
    =
    \mathbf{H}^{l}_{dil}
    +
    \mathcal{D}_{s}
    \left(
        \mathcal{D}_{f}
        \left(
            \mathbf{H}^{l}_{dil}
        \right)
    \right).
\end{equation}
This structured approximation avoids constructing a dense joint graph or applying full attention over all node-modality pairs while retaining the ability to model both spatial dependencies and cross-modality relationships.

\begin{table*}[h]
  \centering
  \caption{Summary of experimental datasets.}
  \label{tab:datasets}
  \setlength{\tabcolsep}{2pt}
  \renewcommand{\arraystretch}{1.15}
  \begin{tabular*}{\textwidth}{@{\extracolsep{\fill}} lccccc}
    \toprule
    \textbf{Attribute} & \textbf{NJRC} & \textbf{SZparking} & \textbf{NYC} & \textbf{MetroFlow} & \textbf{PEMS03} \\
    \midrule
    \textbf{Scenario} & Highway & Parking & Urban Mobility & Metro & Highway \\
    \textbf{Nodes ($N$)} & 36 & 189 & 98 & 302 & 358 \\
    \textbf{Resolution} & 5/10/15 min & 5/10/15 min & 30 min & 10 min & 5 min \\
    \textbf{Duration} 
    & \makecell[c]{2024/9/7--2025/10/14} 
    & \makecell[c]{2024/11/1--2024/11/30} 
    & \makecell[c]{2016/4/1--2016/6/30} 
    & \makecell[c]{2017/5/1--2017/8/30} 
    & \makecell[c]{2018/9/1--2018/11/30} \\
    \textbf{Modalities} 
    & \makecell[c]{Total + \\\{ICEV, NEV\} + \\\{PV, Truck, SV\} + \\\{ETC, CPC\} + \\\{Local, Nonlocal\}} 
    & \makecell[c]{\{In, Out\} $\times$ \\ \{\{Total, ICEV, NEV\} + \\\{Local, Nonlocal\}\}} 
    & \makecell[c]{\{Bike, Taxi\} $\times$ \\ \{Total, In, Out\}} 
    & \makecell[c]{\{In, Out\} $\times$ \\ \{Total, C, HBO, NHB\}} 
    & \makecell[c]{Traffic Flow} \\
    \textbf{Exogenous} & Weather, Holiday & Weather, Holiday & Holiday & Weather, Holiday & Holiday \\
    \textbf{Split} & 237 : 83 : 83 & 19 : 4 : 7 & 6 : 2 : 2 & 6 : 2 : 2 & 6 : 2 : 2 \\
    \bottomrule
  \end{tabular*}
\end{table*}

\section{Experiments}
\label{sec:experiments}
In this section, we conduct extensive experiments to evaluate DSFNet for MoSTF, comparing it with state-of-the-art spatio-temporal models, assessing the contributions of DDIBlock and EGDCBlock via ablation studies, and analyzing its computational efficiency and scalability.

\subsection{Datasets}
We validate the effectiveness of DSFNet on five representative real-world flow datasets covering diverse transportation scenarios, including highways, urban parking, and public transit, as shown in Table~\ref{tab:datasets}. Beyond the \textbf{PEMS03}, we introduce two proprietary multi-modality datasets enriched with exogenous variables (\textbf{NJRC} and \textbf{SZparking}) and two public multi-modality datasets (\textbf{NYC} \cite{deng2024multi} and \textbf{MetroFlow} \cite{sun2025human}). Specifically, \textbf{NJRC} (ETC gantries, Nanjing Ring Expressway) distinguishes flows by powertrain, vehicle class, transaction type, and plate location; \textbf{SZparking} (parking gate data, Shenzhen) differentiates inflow/outflow, powertrain, and plate location; \textbf{NYC} separates bike and taxi flows; and \textbf{MetroFlow} divides passenger flows by commuting purpose.

\subsection{Experimental Settings}
All experiments are implemented using PyTorch 2.5.0 with CUDA 12.4 on an NVIDIA RTX 4090D GPU within the BasicTS framework \cite{shao2024exploring}. Models are trained with the Adam optimizer at an initial learning rate of 0.002. For key modules, the hidden dimension is selected from {8, 16, 32, 64, 128}, with 64 performing best, and the number of layers $L$ is set to 8. Both the dilation factor and diffusion steps are set to 2 \cite{wu2019graph}, and all models use a 12-step-input and 12-step-output forecasting setting \cite{gao2024spatial}. 

We compare DSFNet with the following baselines that belong to three categories: (1) STGNN-based models: \textbf{DCRNN}  \cite{li2018dcrnn_traffic}, \textbf{STGCN} \cite{yu2018spatio}, \textbf{GWNet} \cite{wu2019graph}, \textbf{MTGNN} \cite{wu2020connecting}, and \textbf{STPGNN} \cite{kong2024spatio}; (2) Transformer-based models: \textbf{Autoformer} \cite{wu2021autoformer}, \textbf{Crossformer} \cite{zhang2023crossformer}, \textbf{iTransformer} \cite{liu2024itransformer}, \textbf{TimeXer} \cite{wang2024timexer}, and \textbf{PatchTST} \cite{nie2023a}; and (3) Linear or hybrid models: \textbf{DLinear} \cite{zeng2023transformers} and \textbf{TiDE} \cite{das2023longterm}. We use Mean Absolute Error (MAE), Mean Absolute Percentage Error (MAPE), and Root Mean Squared Error (RMSE) as evaluation metrics, averaging over all nodes, variables, and prediction horizons.

%%%%%%%%%%%%%%%%%%%%%%%%%%%%%%%%%%%%%%%%%%%%%%%%%%%%%%%%%%%%%%%%%%%%%%%%%%%%%%%%%%%%%%%%%%%%%%%%%%%%%%%%%%%%%%%%%%%%%%%%%%%%%%%%
\begin{table*}[h]
  \centering
  \small
  \caption{Performance comparison across representative datasets.}
  \label{tab:main_comparison}
  \setlength{\tabcolsep}{0.8pt}
  \renewcommand{\arraystretch}{1.05}
  \begin{tabular*}{\textwidth}{@{\extracolsep{\fill}} lccccccccccccccc}
    \toprule
    \multirow{2}{*}{\textbf{Model}} 
    & \multicolumn{3}{c}{\makecell[c]{\textbf{NJRC} \textbf{10min}}}
    & \multicolumn{3}{c}{\makecell[c]{\textbf{SZparking} \textbf{10min}}}
    & \multicolumn{3}{c}{\makecell[c]{\textbf{NYC} \textbf{30min}}}
    & \multicolumn{3}{c}{\makecell[c]{\textbf{MetroFlow} \textbf{10min}}}
    & \multicolumn{3}{c}{\makecell[c]{\textbf{PEMS03} \textbf{5min}}} \\
    \cmidrule(lr){2-4}
    \cmidrule(lr){5-7}
    \cmidrule(lr){8-10}
    \cmidrule(lr){11-13}
    \cmidrule(lr){14-16}
    & \textbf{MAE} & \textbf{MAPE} & \textbf{RMSE}
    & \textbf{MAE} & \textbf{MAPE} & \textbf{RMSE}
    & \textbf{MAE} & \textbf{MAPE} & \textbf{RMSE}
    & \textbf{MAE} & \textbf{MAPE} & \textbf{RMSE}
    & \textbf{MAE} & \textbf{MAPE} & \textbf{RMSE} \\
    \midrule
    DCRNN 
    & 9.14 & 26.5\% & 17.09
    & 2.33 & 59.7\% & 5.23
    & 13.56 & 51.9\% & 27.35
    & 17.84 & \underline{31.5\%} & 41.23
    & 16.96 & 16.8\% & 27.57 \\
    
    STGCN 
    & 19.27 & 68.9\% & 36.16
    & 3.36 & 86.0\% & 7.44
    & 32.42 & 171.3\% & 61.09
    & 54.74 & 122.8\% & 137.04
    & 27.29 & 54.6\% & 43.40 \\
    
    GWNet 
    & \underline{8.71} & \underline{25.9\%} & \underline{16.16}
    & \underline{1.87} & \underline{51.5\%} & \underline{3.36}
    & \underline{11.87} & \underline{43.9\%} & \underline{22.83}
    & \underline{16.48} & 32.4\% & \underline{36.22}
    & \textbf{15.04} & \underline{14.7\%} & \textbf{24.23$^{*}$} \\
    
    MTGNN 
    & 9.69 & 30.1\% & 18.41
    & 2.88 & 71.0\% & 7.55
    & 30.98 & 148.3\% & 59.35
    & 42.58 & 93.5\% & 120.17
    & 24.80 & 23.4\% & 39.15 \\
    
    STPGNN 
    & 18.13 & 61.9\% & 33.97
    & 2.88 & 72.8\% & 6.55
    & 23.37 & 118.6\% & 43.67
    & 40.18 & 79.5\% & 106.84
    & 51.50 & 112.7\% & 70.19 \\
    
    \midrule
    Autoformer 
    & 10.26 & 34.5\% & 18.15
    & 2.06 & 54.0\% & 4.09
    & 13.24 & 60.3\% & 24.89
    & 19.46 & 39.5\% & 43.61
    & 16.88 & 18.6\% & 26.37 \\
    
    Crossformer 
    & 10.48 & 37.4\% & 17.72
    & 2.22 & 56.1\% & 5.10
    & 15.11 & 78.8\% & 26.74
    & 34.64 & 67.2\% & 87.32
    & \underline{15.09} & \textbf{14.5\%} & 25.58 \\
    
    iTransformer 
    & 9.68 & 30.8\% & 17.78
    & 2.23 & 54.1\% & 4.97
    & 19.50 & 81.9\% & 39.57
    & 28.50 & 47.9\% & 76.36
    & 17.42 & 16.6\% & 27.85 \\
    
    TimeXer 
    & 10.08 & 30.6\% & 19.01
    & 2.05 & \underline{51.5\%} & 4.41
    & 12.97 & 55.9\% & 25.50
    & 18.23 & 34.7\% & 43.11
    & 16.07 & 15.3\% & 26.07 \\
    
    PatchTST 
    & 14.26 & 39.0\% & 27.40
    & 2.58 & 62.9\% & 5.82
    & 26.87 & 108.7\% & 53.77
    & 38.61 & 64.1\% & 103.79
    & 21.52 & 19.6\% & 34.37 \\
    
    \midrule
    DLinear 
    & 14.57 & 43.3\% & 28.35
    & 2.53 & 63.2\% & 5.70
    & 25.09 & 96.8\% & 48.43
    & 38.68 & 68.5\% & 110.42
    & 21.23 & 20.6\% & 34.16 \\
    
    TiDE 
    & 14.42 & 45.2\% & 26.93
    & 2.51 & 68.2\% & 5.21
    & 24.44 & 127.1\% & 46.71
    & 38.76 & 83.1\% & 99.51
    & 18.75 & 19.1\% & 29.37 \\
    
    \midrule
    DSFNet (ours) 
    & \textbf{8.43$^{*}$} & \textbf{25.4\%$^{*}$} & \textbf{15.18$^{*}$}
    & \textbf{1.68$^{*}$} & \textbf{43.4\%$^{*}$} & \textbf{2.95$^{*}$}
    & \textbf{10.81$^{*}$} & \textbf{40.5\%$^{*}$} & \textbf{21.14$^{*}$}
    & \textbf{15.25$^{*}$} & \textbf{29.1\%$^{*}$} & \textbf{33.67$^{*}$}
    & 15.72 & 15.6\% & \underline{25.37} \\
    \bottomrule
  \end{tabular*}

  \vspace{2pt}
  \begin{minipage}{\textwidth}
  \normalsize
  \textbf{Note:} Lower values indicate better performance. 
  \textbf{Bold}: Best, \underline{Underline}: Second best. 
  $^*$ indicates statistically significant improvement ($p<0.05$).
  \end{minipage}
\end{table*}

\begin{table*}[h]
  \centering
  \small
  \caption{Performance comparison under different sampling intervals.}
  \label{tab:sampling_comparison}
  \setlength{\tabcolsep}{0.8pt}
  \renewcommand{\arraystretch}{1.05}

  \begin{tabular*}{\textwidth}{@{\extracolsep{\fill}} llccccccccc}
    \toprule
    \multirow{2}{*}{\textbf{Dataset}}
    & \multirow{2}{*}{\textbf{Model}}
    & \multicolumn{3}{c}{\textbf{5 min}}
    & \multicolumn{3}{c}{\textbf{10 min}}
    & \multicolumn{3}{c}{\textbf{15 min}} \\
    \cmidrule(lr){3-5}
    \cmidrule(lr){6-8}
    \cmidrule(lr){9-11}
    & & \textbf{MAE} & \textbf{MAPE} & \textbf{RMSE}
      & \textbf{MAE} & \textbf{MAPE} & \textbf{RMSE}
      & \textbf{MAE} & \textbf{MAPE} & \textbf{RMSE} \\
    \midrule

    \multirow{13}{*}{\textbf{NJRC}}
    & DCRNN 
    & 4.91 & \textcolor{red}{29.5\%} & 7.96
    & 9.14 & 26.5\% & 17.09
    & 14.00 & \textcolor{red}{25.8\%} & 28.55 \\

    & STGCN 
    & 7.44 & \textcolor{red}{46.0\%} & 12.85
    & 19.27 & 68.9\% & 36.16
    & 27.17 & 74.6\% & 48.13 \\

    & GWNet 
    & \textbf{4.85} & \textbf{29.4\%} & \textbf{7.84}
    & \underline{8.71} & \underline{25.9\%} & \underline{16.16}
    & \underline{12.82} & \textcolor{red}{\underline{24.9\%}} & \underline{25.55} \\

    & MTGNN 
    & 5.08 & 31.0\% & 8.32
    & 9.69 & \textcolor{red}{30.1\%} & 18.41
    & 14.87 & 32.2\% & 29.49 \\

    & STPGNN 
    & 8.33 & \textcolor{red}{58.6\%} & 14.57
    & 18.13 & 61.9\% & 33.97
    & 28.34 & 78.2\% & 52.65 \\

    & Autoformer 
    & 5.26 & \textcolor{red}{33.4\%} & 8.54
    & 10.26 & 34.5\% & 18.15
    & 18.59 & 45.2\% & 34.61 \\

    & Crossformer 
    & 5.77 & 38.8\% & 8.99
    & 10.48 & \textcolor{red}{37.4\%} & 17.72
    & 15.37 & 37.5\% & 26.82 \\

    & iTransformer 
    & 5.28 & 32.6\% & 8.67
    & 9.68 & \textcolor{red}{30.8\%} & 17.78
    & 15.18 & 32.6\% & 30.01 \\

    & TimeXer 
    & 5.26 & 32.3\% & 8.63
    & 10.08 & \textcolor{red}{30.6\%} & 19.01
    & 15.66 & 32.0\% & 31.18 \\

    & PatchTST 
    & 6.80 & \textcolor{red}{38.3\%} & 11.83
    & 14.26 & 39.0\% & 27.40
    & 24.34 & 44.4\% & 47.77 \\

    & DLinear 
    & 6.67 & \textcolor{red}{39.8\%} & 11.62
    & 14.57 & 43.3\% & 28.35
    & 25.18 & 50.7\% & 49.18 \\

    & TiDE 
    & 6.13 & \textcolor{red}{39.0\%} & 10.22
    & 14.42 & 45.2\% & 26.93
    & 26.53 & 59.7\% & 49.33 \\

    \cmidrule(lr){2-11}
    & DSFNet (ours) 
    & \underline{4.86} & \underline{29.5\%} & \underline{7.85}
    & \textbf{8.43$^{*}$} & \textbf{25.4\%$^{*}$} & \textbf{15.18$^{*}$}
    & \textbf{12.63$^{*}$} & \textcolor{red}{\textbf{24.4\%$^{*}$}} & \textbf{24.77$^{*}$} \\

    \midrule

    \multirow{13}{*}{\textbf{SZparking}}
    & DCRNN 
    & 1.52 & 56.8\% & 2.86
    & 2.33 & 59.7\% & 5.23
    & 3.11 & \textcolor{red}{55.5\%} & 6.98 \\

    & STGCN 
    & 2.20 & 75.2\% & 4.39
    & 3.36 & 86.0\% & 7.44
    & 5.05 & \textcolor{red}{67.2\%} & 11.63 \\

    & GWNet 
    & \underline{1.25} & \textcolor{red}{\underline{43.0\%}} & \underline{2.15}
    & \underline{1.87} & \underline{51.5\%} & \underline{3.36}
    & \underline{2.63} & \underline{50.3\%} & \underline{5.17} \\

    & MTGNN 
    & 1.63 & \textcolor{red}{58.2\%} & 4.14
    & 2.88 & 71.0\% & 7.55
    & 3.95 & 60.0\% & 9.17 \\

    & STPGNN 
    & 1.78 & 60.6\% & 3.61
    & 2.88 & 72.8\% & 6.55
    & 3.91 & \textcolor{red}{54.5\%} & 9.32 \\

    & Autoformer 
    & 1.40 & \textcolor{red}{49.2\%} & 2.47
    & 2.06 & 54.0\% & 4.09
    & 3.07 & 57.6\% & 6.29 \\

    & Crossformer 
    & 1.48 & \textcolor{red}{51.3\%} & 2.91
    & 2.22 & 56.1\% & 5.10
    & 3.36 & 54.4\% & 8.01 \\

    & iTransformer 
    & 1.39 & \textcolor{red}{43.9\%} & 2.62
    & 2.23 & 54.1\% & 4.97
    & 3.25 & 57.6\% & 7.03 \\

    & TimeXer 
    & 1.34 & \textcolor{red}{44.7\%} & 2.41
    & 2.05 & 51.5\% & 4.41
    & 2.85 & 53.2\% & 5.87 \\

    & PatchTST 
    & 1.60 & \textcolor{red}{57.8\%} & 3.01
    & 2.58 & 62.9\% & 5.82
    & 3.56 & 64.1\% & 7.71 \\

    & DLinear 
    & 1.56 & \textcolor{red}{56.3\%} & 2.91
    & 2.53 & 63.2\% & 5.70
    & 3.48 & 57.4\% & 7.79 \\

    & TiDE 
    & 1.63 & \textcolor{red}{62.4\%} & 2.73
    & 2.51 & 68.2\% & 5.21
    & 3.44 & 73.7\% & 6.72 \\

    \cmidrule(lr){2-11}
    & DSFNet (ours) 
    & \textbf{1.21$^{*}$} & \textcolor{red}{\textbf{39.3\%$^{*}$}} & \textbf{2.11$^{*}$}
    & \textbf{1.68$^{*}$} & \textbf{43.4\%$^{*}$} & \textbf{2.95$^{*}$}
    & \textbf{2.37$^{*}$} & \textbf{44.4\%$^{*}$} & \textbf{4.28$^{*}$} \\

    \bottomrule
  \end{tabular*}

  \vspace{2pt}
  \begin{minipage}{\textwidth}
  \normalsize
    \textbf{Note:} Lower values indicate better performance. 
    \textbf{Bold}: Best, \underline{Underline}: Second best. 
    \textcolor{red}{Red}: Best MAPE across different sampling intervals for each model. 
    $^*$ indicates statistically significant improvement ($p<0.05$).
  \end{minipage}
\end{table*}

\begin{table*}[h]
\centering
\caption{MAE comparison at different forecast steps across five datasets.}
\label{tab:mae_steps_all}
\newcolumntype{Y}{>{\centering\arraybackslash}m{0.82cm}}
\setlength{\tabcolsep}{1pt}
\renewcommand{\arraystretch}{1.0} 

\setlength{\aboverulesep}{0pt}
\setlength{\belowrulesep}{0pt}
\setlength{\tabcolsep}{0pt}
\resizebox{\linewidth}{!}{%
\scriptsize
\begin{tabular}{l | *{4}{Y} | *{4}{Y} | *{4}{Y} | *{4}{Y} | *{4}{Y}}
\toprule
\multirow{2}{*}{\textbf{Model}} &
\multicolumn{4}{c|}{\textbf{NJRC 10min}} &
\multicolumn{4}{c|}{\textbf{SZparking 10min}} &
\multicolumn{4}{c|}{\textbf{NYC 30min}} &
\multicolumn{4}{c|}{\textbf{MetroFlow 10min}} &
\multicolumn{4}{c}{\textbf{PEMS03 5min}} \\
\cmidrule(lr){2-5} \cmidrule(lr){6-9} \cmidrule(lr){10-13} \cmidrule(lr){14-17} \cmidrule(lr){18-21}
& \textbf{S3} & \textbf{S6} & \textbf{S9} & \textbf{S12}
& \textbf{S3} & \textbf{S6} & \textbf{S9} & \textbf{S12}
& \textbf{S3} & \textbf{S6} & \textbf{S9} & \textbf{S12}
& \textbf{S3} & \textbf{S6} & \textbf{S9} & \textbf{S12}
& \textbf{S3} & \textbf{S6} & \textbf{S9} & \textbf{S12} \\
\midrule
DCRNN      & 7.44 & 8.84 & 10.42 & 11.77 
           & 2.15 & 2.30 & 2.45 & 2.61 
           & 11.02 & 13.56 & 15.61 & 17.10
           & \underline{14.45} & 16.95 & 20.10 & 23.67
           & 14.67 & 16.75 & 18.76 & 20.74 \\
GWNet      & \underline{7.40} & \underline{8.46} & \underline{9.70} & \underline{10.73} 
           & \underline{1.80} & \underline{1.87} & \underline{1.92} & \underline{1.97} 
           & \underline{10.66} & \underline{11.99} & \underline{12.92} & \underline{13.92}
           & 14.87 & \underline{16.24} & \underline{17.57} & \underline{19.24}
           & \textbf{13.62} & \textbf{15.00} & \textbf{16.19} & \textbf{17.20} \\
Autoformer & 8.62 & 9.82 & 11.39 & 12.83 
           & 1.93 & 2.03 & 2.11 & 2.36 
           & 12.54 & 13.29 & 13.86 & 14.77
           & 17.12 & 18.32 & 20.85 & 25.22
           & 15.10 & 16.48 & 18.15 & 20.06 \\
TimeXer    & 7.84 & 9.57 & 11.69 & 13.66 
           & 1.88 & 2.04 & 2.18 & 2.31 
           & 10.90 & 13.20 & 14.63 & 15.80
           & 15.02 & 17.74 & 20.45 & 23.22
           & \underline{13.91} & 15.83 & 17.67 & 19.55 \\
TiDE       & 10.44 & 13.63 & 17.30 & 20.06 
           & 2.23 & 2.49 & 2.65 & 2.93 
           & 18.79 & 25.34 & 28.69 & 31.72 
           & 27.30 & 39.81 & 48.12 & 54.21 
           & 15.24 & 18.15 & 21.24 & 24.85 \\
\midrule
\textbf{DSFNet} &
% NJRC
\makecell[c]{\textbf{7.33}\\\textbf{\tiny (0.95\%)}} &
\makecell[c]{\textbf{8.25}\\\textbf{\tiny (2.48\%)}} &
\makecell[c]{\textbf{9.23}\\\textbf{\tiny (4.85\%)}} &
\makecell[c]{\textbf{10.10}\\\textbf{\tiny (5.87\%)}} &
% SZparking
\makecell[c]{\textbf{1.65}\\\textbf{\tiny (8.33\%)}} &
\makecell[c]{\textbf{1.67}\\\textbf{\tiny (10.7\%)}} &
\makecell[c]{\textbf{1.70}\\\textbf{\tiny (11.5\%)}} &
\makecell[c]{\textbf{1.72}\\\textbf{\tiny (12.7\%)}} &
% NYC
\makecell[c]{\textbf{9.71}\\\textbf{\tiny (8.95\%)}} &
\makecell[c]{\textbf{10.82}\\\textbf{\tiny (9.78\%)}} &
\makecell[c]{\textbf{11.73}\\\textbf{\tiny (9.19\%)}} &
\makecell[c]{\textbf{12.67}\\\textbf{\tiny (8.99\%)}} &
% MetroFlow
\makecell[c]{\textbf{13.71}\\\textbf{\tiny (5.14\%)}} &
\makecell[c]{\textbf{14.87}\\\textbf{\tiny (8.41\%)}} &
\makecell[c]{\textbf{16.26}\\\textbf{\tiny (7.50\%)}} &
\makecell[c]{\textbf{18.16}\\\textbf{\tiny (5.64\%)}} &
% PEMS03
13.92 & \underline{15.62} & \underline{17.12} & \underline{18.51} \\
\bottomrule
\end{tabular}%
}
\begin{minipage}{\linewidth}
\footnotesize
\textbf{Note:} \textbf{Bold} and \underline{underlined} fonts indicate the best and second-best results. (Number) denotes the relative improvement over the second-best baseline. S3, S6, S9, S12 denote forecast steps 3, 6, 9, 12 respectively.
\end{minipage}
\end{table*}
%%%%%%%%%%%%%%%%%%%%%%%%%%%%%%%%%%%%%%%%%%%%%%%%%%%%%%%%%%%%%%%%%%%%%%%%%%%%%%%%%%%%%%%%%%%%%%%%%%%%%%%%%%%%%%%%%%%%%%%%%%%%%%%%

\subsection{Overall Performance Comparison}

All results are averaged over ten independent runs to ensure statistical reliability. 
We analyze the forecasting performance from four perspectives: overall accuracy across representative datasets, the influence of sampling intervals, modality-wise forecasting performance, and horizon-wise prediction stability.

\paragraph{Overall performance across representative datasets.}
Table~\ref{tab:main_comparison} reports the overall comparison on five representative datasets. 
DSFNet achieves the best performance on all four multi-modality datasets, including NJRC, SZparking, NYC, and MetroFlow. 
In terms of MAE, DSFNet improves over the second-best baseline by 3.21\% on NJRC, 10.16\% on SZparking, 8.93\% on NYC, and 7.46\% on MetroFlow. 
These results indicate that DSFNet is not tailored to a specific traffic scenario, but generalizes across highway traffic, parking systems, urban mobility, and metro passenger flows.

The comparison also reveals clear differences among model families. 
STGNN-based models, especially GWNet, remain strong competitors because adaptive graph learning provides useful spatial inductive bias. 
However, their shared graph propagation is less effective when different traffic modalities exhibit heterogeneous spatial patterns. 
Transformer-based models can capture cross-variable dependencies, but their shared attention representations may entangle modality-specific dynamics and lack explicit transportation-network inductive bias. 
Linear and hybrid models, such as DLinear and TiDE, are generally less competitive on MoSTF datasets, suggesting that simplified temporal mappings are insufficient for nonlinear modality interactions.

The single-modality PEMS03 dataset provides an informative contrast. 
DSFNet is slightly inferior to GWNet on PEMS03, where GWNet achieves the best MAE and RMSE. 
This result is reasonable because PEMS03 mainly involves conventional single-variable traffic-flow forecasting, where graph propagation alone is already highly effective. 
In contrast, the consistent gains on the four multi-modality datasets support the core motivation of this paper: DSFNet is particularly beneficial when traffic forecasting requires joint modeling of spatial dependencies and cross-modality relationships.

\paragraph{Performance under different sampling intervals.}
Table~\ref{tab:sampling_comparison} investigates how temporal granularity affects forecasting performance on NJRC and SZparking. 
The results show that sampling interval is not merely a preprocessing choice, but directly changes the prediction difficulty. 
On NJRC, DSFNet is almost tied with GWNet under the 5min setting, with only a 0.01 MAE difference. 
This suggests that very fine-grained highway data contain stronger short-term fluctuations, where adaptive graph propagation remains highly competitive. 
When the interval increases to 10min and 15min, DSFNet becomes the best-performing model, indicating that moderate temporal aggregation makes stable spatial-modality patterns easier to capture.

The trend is more pronounced on SZparking. 
DSFNet ranks first under all three sampling intervals and achieves particularly large gains at 10min and 15min. 
For example, at 10min, DSFNet reduces MAE by 10.16\% and MAPE by 15.73\% over the second-best baseline. 
At 15min, it further reduces RMSE by 17.21\%. 
These improvements suggest that parking flows contain strong modality coupling, such as inflow--outflow relationships and NEV--ICEV heterogeneity. 
The relatively stable MAPE of DSFNet across 5min, 10min, and 15min also shows that the model maintains robust relative prediction accuracy under different temporal granularities.

Overall, the sampling-interval analysis suggests that DSFNet is robust to changes in temporal resolution. 
Its advantage becomes more evident when the data contain meaningful cross-modality structures rather than only high-frequency local fluctuations.

\begin{figure}[h]
    \centering
    \includegraphics[width=\linewidth]{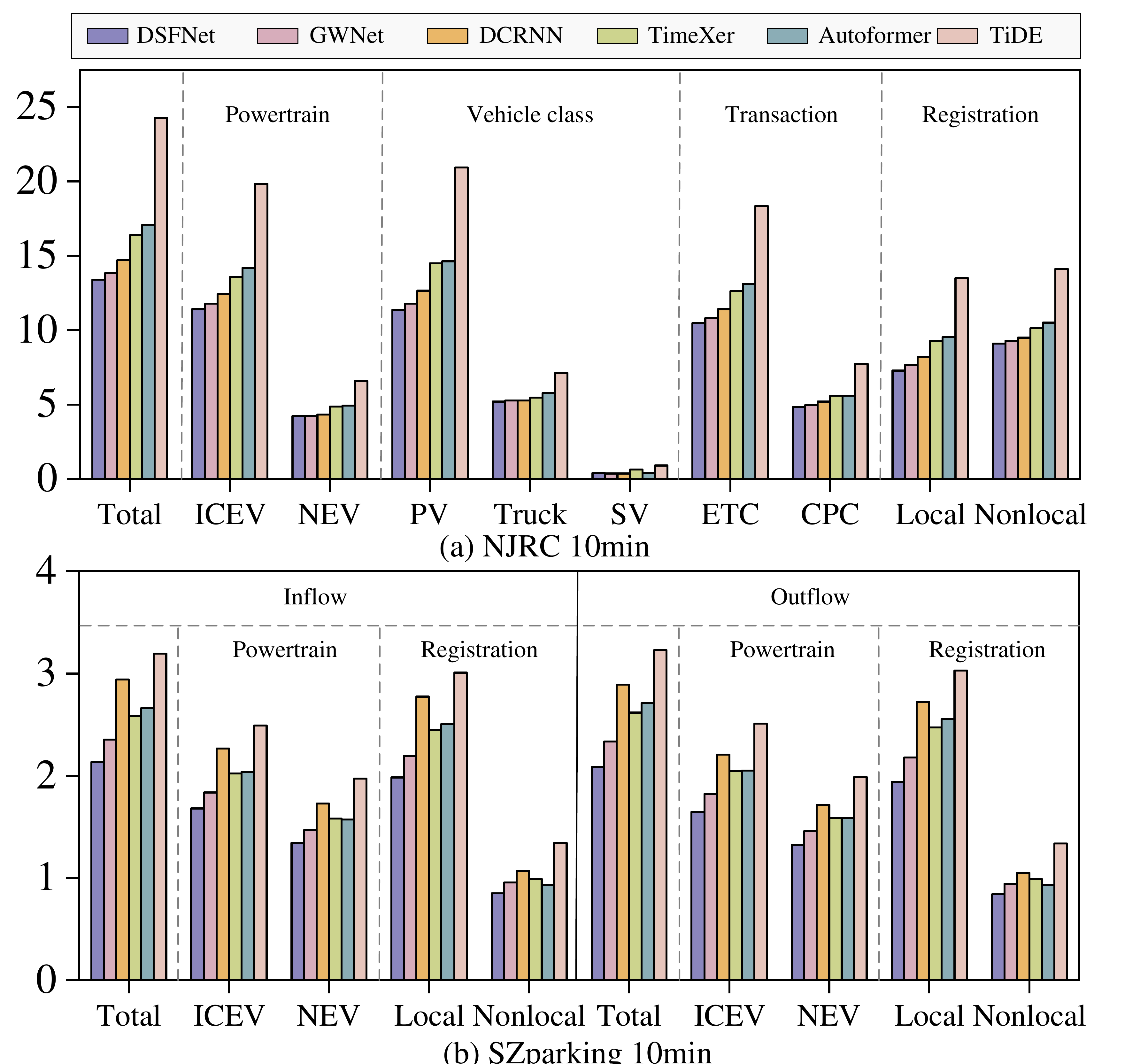}
    \caption{MAE comparison across different multi-modality components: (a) NJRC 10min and (b) SZparking 10min.}
    \label{fig:FineG_MAE}
\end{figure}

\paragraph{Modality-wise performance.}
Fig.~\ref{fig:FineG_MAE} presents modality-wise MAE comparisons on NJRC 10min and SZparking 10min. 
DSFNet achieves the best performance across all modalities on SZparking and remains superior for most modalities on NJRC. 
This result directly supports the necessity of modality-aware modeling in MoSTF. 
Different traffic modalities are not independent: parking inflow and outflow are coupled by parking duration, while NEV and ICEV flows may share common demand patterns but differ in charging-related behavior and facility preference.

Compared with baselines that mainly rely on shared graphs, shared attention layers, or independent temporal modeling, DSFNet explicitly introduces feature-domain spectral filtering before spatial-domain filtering. 
This design allows related modalities to exchange information while preserving modality-specific dynamics. 
The slight degradation on the special-vehicle modality in NJRC also provides useful insight: sparse and event-driven modalities may require additional context or event information beyond historical traffic patterns. 
Nevertheless, the overall modality-wise results validate that DSFNet can effectively exploit cross-modality dependencies in most practical MoSTF settings.

\paragraph{Horizon-wise prediction performance.}
Table~\ref{tab:mae_steps_all} reports MAE at forecast steps S3, S6, S9, and S12. 
As expected, prediction errors increase as the horizon becomes longer. 
However, DSFNet shows stronger long-horizon stability on multi-modality datasets. 
On NJRC, the relative improvement over the second-best baseline increases from 0.95\% at S3 to 5.87\% at S12. 
On SZparking, the improvement grows from 8.33\% to 12.7\% over the same horizon range. 
This trend indicates that the proposed dual-domain interaction becomes more valuable when forecasting becomes more difficult and error accumulation becomes more severe.

On NYC and MetroFlow, DSFNet also maintains stable gains across horizons, with improvements around 9\% on NYC and 5--8\% on MetroFlow. 
These results suggest that DSFNet not only improves average accuracy but also provides more reliable multi-step forecasting for heterogeneous mobility systems. 
On PEMS03, GWNet remains the best model across horizons, while DSFNet achieves competitive second-best results at medium and long horizons. 
This again confirms that the main advantage of DSFNet lies in MoSTF scenarios, where cross-modality interaction and heterogeneous spatial dependency jointly affect future traffic states.

\subsection{Ablation Study}

\paragraph{Effects of DDIBlock}
To evaluate the contribution of the Dual-Domain Interaction Block (DDIBlock), we construct four variants: 
(a) \textbf{w/o DDIBlock}, where the entire DDIBlock is replaced with a standard CNN; 
(b) \textbf{GCN}, where DDIBlock is replaced with a standard graph convolutional layer; 
(c) \textbf{w/o Spatial Operator}, where the spatial-domain spectral operator is removed; and 
(d) \textbf{w/o Feature Operator}, where the feature-domain spectral operator is removed.

\begin{figure}[h]
    \centering
    \includegraphics[width=\linewidth]{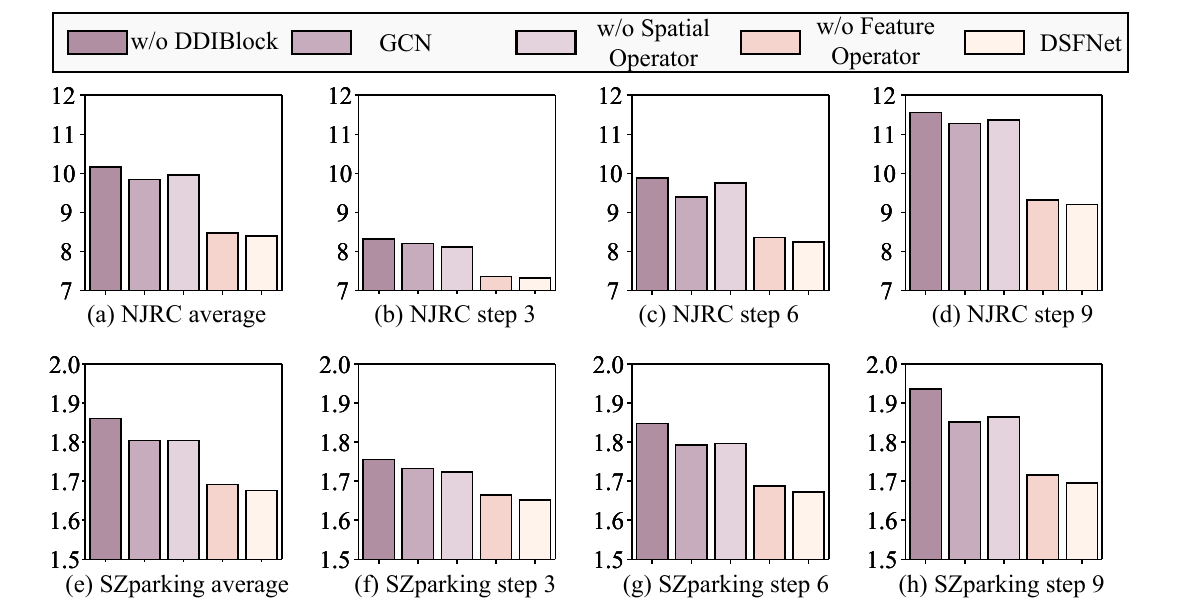}
    \caption{MAE comparison of DDIBlock ablations under the 10min setting.}
    \label{fig:Figure_DDIBlock_Ablation_MAE}
\end{figure}

As shown in Fig.~\ref{fig:Figure_DDIBlock_Ablation_MAE}, all variants lead to performance degradation, confirming the necessity of dual-domain interaction modeling. 
Replacing DDIBlock with a standard CNN causes a clear accuracy drop, indicating that local convolution is insufficient to capture global spatial-modality dependencies in MoSTF. 
The GCN variant performs better than the CNN replacement in some cases, showing that graph-based spatial inductive bias is useful for traffic forecasting. 
However, it still underperforms the complete DSFNet, suggesting that spatial message passing alone cannot fully characterize the interactions among heterogeneous traffic modalities. 
This result is consistent with our motivation that MoSTF requires not only entity-wise spatial modeling, but also modality-wise interaction modeling.

The comparison between \textbf{w/o Spatial Operator} and \textbf{w/o Feature Operator} further reveals the different roles of the two spectral operators. 
Removing the spatial operator causes a substantial performance decline, demonstrating that spatial dependencies among transportation entities remain fundamental for traffic forecasting. 
Removing the feature operator also degrades accuracy, which validates the importance of cross-modality interaction modeling. 
In particular, traffic modalities such as inflow and outflow, NEV and ICEV, or local and nonlocal flows are correlated but not identical. 
The feature-domain operator enables information exchange among these modalities before spatial-domain filtering, thereby improving modality-aware representation learning. 
Overall, the ablation results verify that the performance gain of DSFNet comes from the structured combination of feature-domain and spatial-domain spectral filtering, rather than from simply increasing model complexity.

\paragraph{Effects of EGDCBlock}
We further evaluate the Exogenous-Gated Dilated Convolution Block (EGDCBlock) using three variants: 
(a) \textbf{w/o EGDCBlock}, where EGDCBlock is replaced with a standard CNN; 
(b) \textbf{w/o SkipGate}, where the skip-gating mechanism is removed; and 
(c) \textbf{w/o Exo}, where exogenous variables are excluded from the gating mechanism.

\begin{figure}[h]
    \centering
    \includegraphics[width=\linewidth]{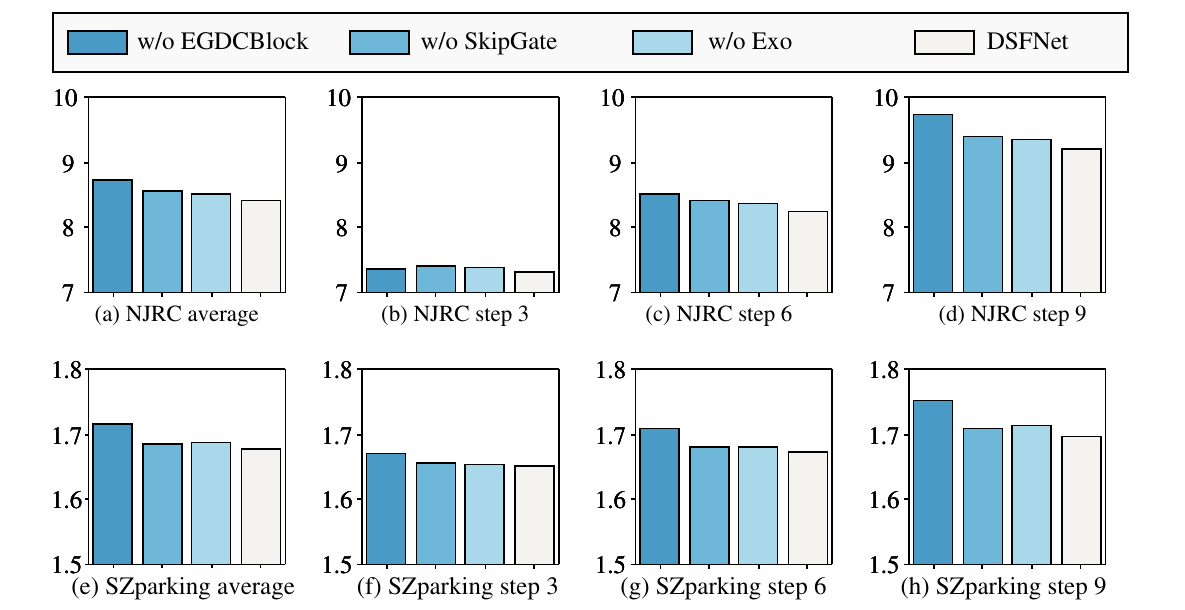}
    \caption{MAE comparison of EGDCBlock ablations under the 10min setting.}
    \label{fig:Figure_EGDCBlock_Ablation_MAE}
\end{figure}

As shown in Fig.~\ref{fig:Figure_EGDCBlock_Ablation_MAE}, removing EGDCBlock leads to a marked performance decline, indicating that simple convolutional temporal modeling cannot fully capture the multi-scale temporal dynamics in MoSTF. 
The dilated causal convolution in EGDCBlock expands the temporal receptive field efficiently, which is important for capturing delayed effects such as parking duration, commuting cycles, and time-dependent traffic demand. 
The degradation of \textbf{w/o SkipGate} further shows that the parameterized skip-gating mechanism is not merely a training trick, but contributes to stable temporal information flow across layers.

The \textbf{w/o Exo} variant also performs worse than the complete DSFNet, demonstrating the value of exogenous-aware temporal modulation. 
External variables, such as weather and holidays, usually do not affect all traffic modalities uniformly. 
For example, holidays may change the balance between local and nonlocal traffic, while weather conditions may influence parking demand and vehicle-type distributions. 
By injecting exogenous embeddings into the gating mechanism, EGDCBlock allows the model to dynamically adjust temporal representations under different external contexts. 
These results support the design motivation in the Introduction: MoSTF requires temporal modeling that is aware of external disturbances rather than relying only on endogenous traffic history.

\begin{figure}[h]
    \centering
    \includegraphics[width=\linewidth]{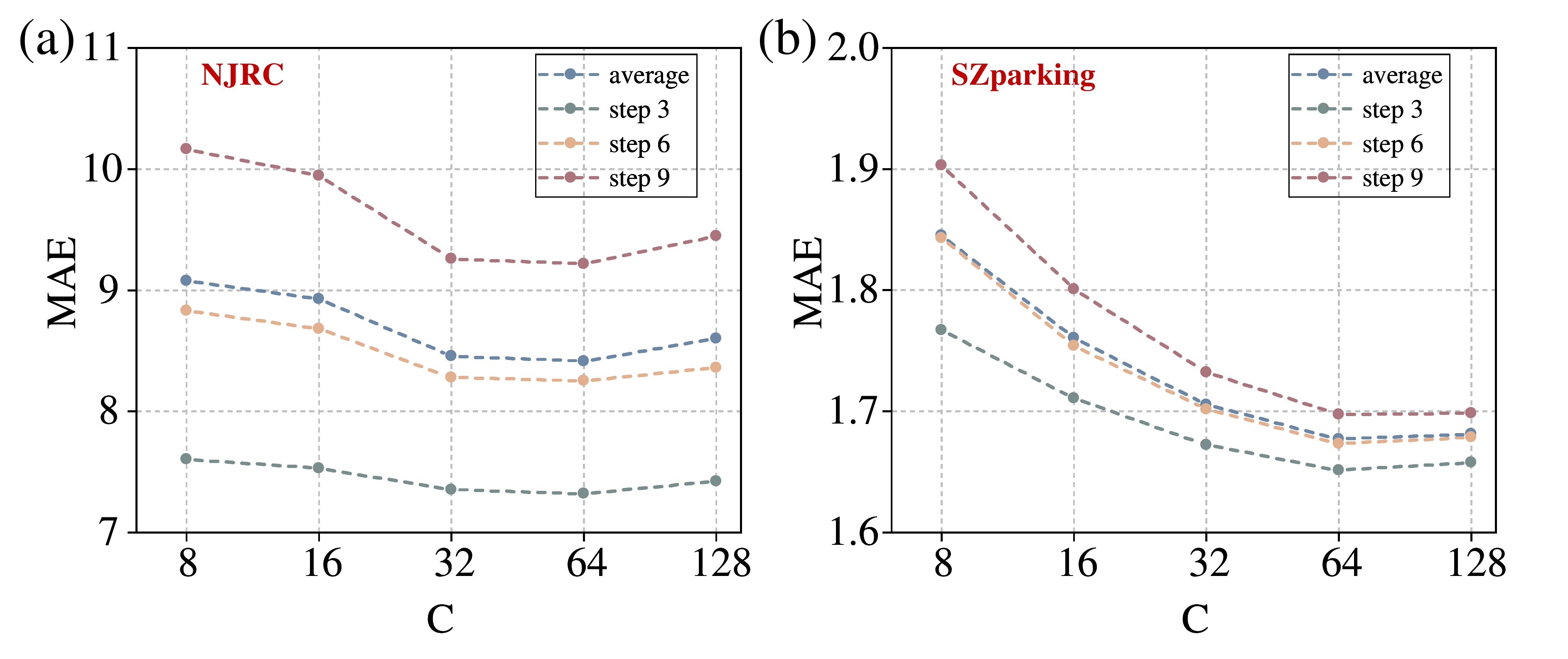}
    \caption{MAE comparison under different hidden dimensions. (a) NJRC 10min and (b) SZparking 10min.}
    \label{fig:Figure_C_Ablation}
\end{figure}

\paragraph{Hyperparameter Study}
We investigate the effect of the hidden dimension $C$ in DSFNet. 
Specifically, we vary $C$ from small to large values and evaluate the corresponding forecasting performance. 
As shown in Fig.~\ref{fig:Figure_C_Ablation}, the performance improves when $C$ increases from a small value, indicating that a sufficiently expressive latent space is necessary to encode temporal, spatial, and modality-related information. 
However, the improvement gradually saturates when $C$ becomes large. 
This suggests that excessively increasing the hidden dimension introduces redundant capacity and may not bring additional benefits for MoSTF. 
Considering both accuracy and computational cost, we set $C=64$ in all experiments.

\subsection{Efficiency and Robustness Studies}

\paragraph{Computational efficiency}
To assess practical deployability, we compare DSFNet with TimeXer in terms of parameter size, training time per epoch, inference latency, and GPU memory usage under the 10min sampling interval on NJRC and SZparking. 
As reported in Table~\ref{tab:eff_10min}, DSFNet is consistently more efficient than TimeXer on both datasets.

On NJRC, DSFNet reduces the number of parameters from 1.69M to 1.23M, corresponding to a 27.2\% reduction. 
It also reduces the training time per epoch from 51.96s to 29.12s and the inference latency from 5.29s to 3.95s, achieving reductions of 44.0\% and 25.3\%, respectively. 
The most significant improvement appears in GPU memory usage, where DSFNet reduces memory consumption from 8.10GB to 2.27GB, saving approximately 72.0\% memory.

The efficiency advantage is even more evident on SZparking. 
DSFNet reduces parameters by 37.0\%, training time by 64.4\%, inference latency by 59.7\%, and memory usage by 82.1\% compared with TimeXer. 
These results demonstrate that the proposed DCT-based dual-domain spectral filtering is computationally efficient. 
Instead of applying dense attention over all node-modality pairs, DSFNet factorizes the interaction into feature-domain and spatial-domain spectral operations. 
This structured approximation avoids the high memory cost of generic attention mechanisms and makes DSFNet more suitable for large-scale transportation deployment.

\begin{table}[h]
\centering
\caption{Efficiency comparison under the 10min setting.}
\label{tab:eff_10min}
\normalsize 
\setlength{\tabcolsep}{3pt}
\renewcommand{\arraystretch}{0.85}
\setlength{\aboverulesep}{2pt} 
\setlength{\belowrulesep}{2pt}
\begin{tabular*}{\linewidth}{@{\extracolsep{\fill}}llcccc}
\toprule
\textbf{Dataset} & \textbf{Model} 
& \textbf{Param $\downarrow$} 
& \textbf{Time $\downarrow$} 
& \textbf{Infer $\downarrow$} 
& \textbf{Mem $\downarrow$} \\
& & (M) & (s/epoch) & (s) & (GB) \\
\midrule
\multirow{2}{*}{NJRC} 
& TimeXer & 1.69 & 51.96 & 5.29 & 8.10 \\
& \textbf{DSFNet} & \textbf{1.23} & \textbf{29.12} & \textbf{3.95} & \textbf{2.27} \\
\midrule
\multirow{2}{*}{SZparking} 
& TimeXer & 2.08 & 31.62 & 3.05 & 8.44 \\
& \textbf{DSFNet} & \textbf{1.31} & \textbf{11.25} & \textbf{1.23} & \textbf{1.51} \\
\bottomrule
\end{tabular*}
\end{table}

\paragraph{Robustness to perturbations}
We further evaluate robustness under common real-world perturbations. 
Following existing missing-data settings \cite{li2013efficient}, we inject Gaussian noise with $\sigma=0.1$ and simulate three missing-input cases with a masking ratio of 0.25, including node missing, time missing, and mixed missing. 
Table~\ref{tab:robustness_noise_missing} compares DSFNet with GWNet, the strongest STGNN baseline in the overall comparison.

DSFNet shows strong robustness under most perturbation settings. 
On SZparking, DSFNet consistently achieves the best MAE and RMSE under all four corruption types, with especially clear gains under time missing and mixed missing. 
This indicates that the proposed dual-domain interaction can exploit complementary information across traffic modalities and parking facilities when part of the temporal observations is unavailable.

For NJRC, DSFNet also remains more robust under Gaussian noise, time missing, and mixed missing. 
The improvement is particularly evident under time missing, where DSFNet reduces MAE from 16.99 to 14.15 and RMSE from 32.63 to 27.01. 
This suggests that feature-domain and spatial-domain interactions help recover missing temporal information by leveraging both modality correlations and spatial dependencies. 
The only exception appears under node missing, where GWNet performs better. 
This is reasonable because NJRC is a highway gantry dataset with strong spatial connectivity; when node-level observations are heavily removed, graph propagation over the remaining spatial structure can provide a stronger inductive bias. 
In contrast, DSFNet relies on ordered spectral representations, which may be more sensitive when the spatial structure itself is severely disrupted.

Overall, the robustness results show that DSFNet is not only accurate under clean observations but also reliable under noisy and incomplete inputs. 
Its advantage is most evident when temporal observations or mixed observations are corrupted, which is consistent with the design of dual-domain spectral filtering for exploiting complementary modality and spatial information.

\begin{table}[h]
\centering
\caption{Robustness evaluation under the 10min setting.}
\label{tab:robustness_noise_missing}
\setlength{\tabcolsep}{1.5pt}
\resizebox{\linewidth}{!}{%
\begin{tabular}{llcccccccc}
\toprule
\multirow{2}{*}{Dataset} & \multirow{2}{*}{Model} &
\multicolumn{2}{c}{\textbf{Gaussian Noise}} &
\multicolumn{2}{c}{\textbf{Node Missing}} &
\multicolumn{2}{c}{\textbf{Time Missing}} &
\multicolumn{2}{c}{\textbf{Mixed Missing}} \\
\cmidrule(lr){3-4}
\cmidrule(lr){5-6}
\cmidrule(lr){7-8}
\cmidrule(lr){9-10}
& & \textbf{MAE} & \textbf{RMSE} 
  & \textbf{MAE} & \textbf{RMSE} 
  & \textbf{MAE} & \textbf{RMSE} 
  & \textbf{MAE} & \textbf{RMSE} \\
\midrule
\multirow{2}{*}{NJRC}
& GWNet  & 8.89 & 16.49 & \textbf{11.63} & \textbf{22.83} & 16.99 & 32.63 & 14.46 & 28.26 \\
& \textbf{DSFNet} & \textbf{8.81} & \textbf{15.99} & 14.51 & 28.15 & \textbf{14.15} & \textbf{27.01} & \textbf{14.13} & \textbf{27.53} \\
\midrule
\multirow{2}{*}{SZparking}
& GWNet  & 1.88 & 3.40 & 2.21 & 4.63 & 2.21 & 4.68 & 2.19 & 4.59 \\
& \textbf{DSFNet} & \textbf{1.77} & \textbf{3.37} & \textbf{2.04} & \textbf{4.48} & \textbf{1.92} & \textbf{3.86} & \textbf{1.96} & \textbf{4.07} \\
\bottomrule
\end{tabular}%
}
\begin{minipage}{\linewidth}
\footnotesize
\textbf{Note:} Gaussian noise is injected with $\sigma=0.1$.
The masking ratio is set to 0.25 for node missing, time missing, and mixed missing.
\end{minipage}
\end{table}

\section{Conclusion and Future Work}
\label{sec:Conclusion and Future Work}

In this paper, we investigated MoSTF, a challenging task that requires jointly modeling temporal evolution, spatial dependencies, modality interactions, and exogenous influences. To address these challenges, we proposed DSFNet, a dual-domain spectral filtering framework for multi-modality traffic forecasting. DSFNet consists of two key modules. First, the Exogenous-Gated Dilated Convolution Block captures multi-scale temporal dependencies and adaptively modulates temporal information flow according to external factors such as weather conditions and holidays. Second, the Dual-Domain Interaction Block performs serial spectral filtering in the feature and spatial domains, enabling efficient approximation of cross-modality interactions and heterogeneous spatial dependencies without explicitly constructing dense node-modality graphs or applying full attention over all node-modality pairs.

Extensive experiments on five real-world transportation datasets demonstrate the effectiveness of DSFNet. The results show that DSFNet achieves state-of-the-art performance on four multi-modality datasets, covering highway traffic, parking systems, urban mobility, and metro passenger flows. The sampling-interval analysis further indicates that DSFNet remains robust under different temporal granularities, especially when stable spatial-modality patterns are observable. Modality-wise and horizon-wise results verify that DSFNet can effectively exploit cross-modality dependencies and maintain reliable long-horizon forecasting performance. In addition, ablation studies confirm the necessity of both dual-domain spectral interaction and exogenous-aware temporal gating. Efficiency and robustness analyses further show that DSFNet achieves favorable computational cost and maintains stable performance under noisy and incomplete observations, supporting its practical applicability to real-world intelligent transportation systems.

Despite these promising results, several directions remain worth exploring. First, although DSFNet uses exogenous variables through a learnable gating mechanism, future work can incorporate richer contextual information, such as traffic incidents, events, charging demand, land-use attributes, and policy changes, to further enhance forecasting under abnormal conditions. Second, the current dual-domain spectral filtering adopts a structured approximation of space-modality interactions. Future studies may investigate adaptive or dynamic spectral bases to better capture time-varying spatial structures and evolving modality relationships. Third, the interpretability of MoSTF models remains an important issue. Future work will explore how Large Language Models (LLMs) and visual analytics can be integrated with DSFNet to provide human-understandable explanations, summarize forecasting patterns, and support decision-making for traffic management, parking operation, and transportation planning. Finally, we will extend DSFNet to broader transportation scenarios and online deployment settings, where streaming data, distribution shifts, and real-time computational constraints need to be jointly considered.

\section*{Ethical Statement}

There are no ethical issues.

% \section*{Acknowledgments}

% \section{References Section}
% You can use a bibliography generated by BibTeX as a .bbl file.
%  BibTeX documentation can be easily obtained at:
%  http://mirror.ctan.org/biblio/bibtex/contrib/doc/
%  The IEEEtran BibTeX style support page is:
%  http://www.michaelshell.org/tex/ieeetran/bibtex/
 
%  % argument is your BibTeX string definitions and bibliography database(s)
% %\bibliography{IEEEabrv,../bib/paper}
% You can manually copy in the resultant .bbl file and set second argument of $\backslash${\tt{begin}} to the number of references
%  (used to reserve space for the reference number labels box).

\bibliographystyle{IEEEtran}
\bibliography{IEEEabrv,references}

\begin{thebibliography}{10}
\providecommand{\url}[1]{#1}
\csname url@samestyle\endcsname
\providecommand{\newblock}{\relax}
\providecommand{\bibinfo}[2]{#2}
\providecommand{\BIBentrySTDinterwordspacing}{\spaceskip=0pt\relax}
\providecommand{\BIBentryALTinterwordstretchfactor}{4}
\providecommand{\BIBentryALTinterwordspacing}{\spaceskip=\fontdimen2\font plus
\BIBentryALTinterwordstretchfactor\fontdimen3\font minus \fontdimen4\font\relax}
\providecommand{\BIBforeignlanguage}[2]{{%
\expandafter\ifx\csname l@#1\endcsname\relax
\typeout{** WARNING: IEEEtran.bst: No hyphenation pattern has been}%
\typeout{** loaded for the language `#1'. Using the pattern for}%
\typeout{** the default language instead.}%
\else
\language=\csname l@#1\endcsname
\fi
#2}}
\providecommand{\BIBdecl}{\relax}
\BIBdecl

\bibitem{YANG2026105413}
\BIBentryALTinterwordspacing
Y.~Yang, H.~Li, and H.~Tu, ``Real-time od connection demand forecasting based on streaming temporal link prediction in dynamic graphs,'' \emph{Transportation Research Part C: Emerging Technologies}, vol. 182, p. 105413, 2026. [Online]. Available: \url{https://www.sciencedirect.com/science/article/pii/S0968090X25004176}
\BIBentrySTDinterwordspacing

\bibitem{zhang2022forecasting}
X.~Zhang, Y.~Sun, F.~Guan, K.~Chen, F.~Witlox, and H.~Huang, ``Forecasting the crowd: An effective and efficient neural network for citywide crowd information prediction at a fine spatio-temporal scale,'' \emph{Transportation Research Part C: Emerging Technologies}, vol. 143, p. 103854, 2022.

\bibitem{deng2024multi}
J.~Deng, R.~Jiang, J.~Zhang, and X.~Song, ``Multi-modality spatio-temporal forecasting via self-supervised learning,'' in \emph{Proceedings of the Thirty-Third International Joint Conference on Artificial Intelligence}, 2024, pp. 2018--2026.

\bibitem{wang2024timexer}
Y.~Wang, H.~Wu, J.~Dong, G.~Qin, H.~Zhang, Y.~Liu, Y.~Qiu, J.~Wang, and M.~Long, ``Timexer: Empowering transformers for time series forecasting with exogenous variables,'' \emph{Advances in Neural Information Processing Systems}, vol.~37, pp. 469--498, 2024.

\bibitem{mozaffari2020joint}
M.~Mozaffari, H.~A. Abyaneh, M.~Jooshaki, and M.~Moeini-Aghtaie, ``Joint expansion planning studies of ev parking lots placement and distribution network,'' \emph{IEEE Transactions on Industrial Informatics}, vol.~16, no.~10, pp. 6455--6465, 2020.

\bibitem{cui2025weather}
Q.~Cui, C.~Huang, K.~Zhang, C.~Gao, T.~Gu, E.~Ke, and P.~Lin, ``Weather effects on highway travel volume: Electric vs. fuel vehicles,'' \emph{Transportation Research Part D: Transport and Environment}, vol. 145, p. 104805, 2025.

\bibitem{li2025analysis}
S.~Li, Y.~Qian, J.~Zeng, and X.~Wei, ``Analysis of the impact of large vehicles in merging areas based on driver characteristics under vehicle-road coordination,'' \emph{Physica A: Statistical Mechanics and its Applications}, vol. 665, p. 130497, 2025.

\bibitem{karaliopoulos2022matching}
M.~Karaliopoulos, O.~Mastakas, and W.~K. Chai, ``Matching supply and demand in online parking reservation platforms,'' \emph{IEEE Transactions on Intelligent Transportation Systems}, vol.~24, no.~3, pp. 3182--3193, 2022.

\bibitem{ijcai2023p491}
J.~Wu, Q.~Qi, J.~Wang, H.~Sun, Z.~Wu, Z.~Zhuang, and J.~Liao, ``Not only pairwise relationships: fine-grained relational modeling for multivariate time series forecasting,'' in \emph{Proceedings of the Thirty-Second International Joint Conference on Artificial Intelligence}, 2023, pp. 4416--4423.

\bibitem{NADI2024104413}
\BIBentryALTinterwordspacing
A.~Nadi, N.~Yorke-Smith, M.~Snelder, J.~{Van Lint}, and L.~Tavasszy, ``Data-driven preference-based routing and scheduling for activity-based freight transport modelling,'' \emph{Transportation Research Part C: Emerging Technologies}, vol. 158, p. 104413, 2024. [Online]. Available: \url{https://www.sciencedirect.com/science/article/pii/S0968090X23004035}
\BIBentrySTDinterwordspacing

\bibitem{wu2019graph}
Z.~Wu, S.~Pan, G.~Long, J.~Jiang, and C.~Zhang, ``Graph wavenet for deep spatial-temporal graph modeling,'' in \emph{Proceedings of the 28th International Joint Conference on Artificial Intelligence}, 2019, pp. 1907--1913.

\bibitem{han2024adaptive}
M.~Han and Q.~Wang, ``Adaptive graph convolution neural differential equation for spatio-temporal time series prediction,'' \emph{IEEE Transactions on Knowledge and Data Engineering}, 2024.

\bibitem{zhang2024predicting}
H.~Zhang, Y.~Xia, S.~Zhong, K.~Wang, Z.~Tong, Q.~Wen, R.~Zimmermann, and Y.~Liang, ``Predicting carpark availability in singapore with cross-domain data: A new dataset and a data-driven approach,'' in \emph{Proceedings of the thirty-third international joint conference on artificial intelligence}, 2024, pp. 7554--7562.

\bibitem{yu2015multi}
F.~Yu and V.~Koltun, ``Multi-scale context aggregation by dilated convolutions,'' \emph{arXiv preprint arXiv:1511.07122}, 2015.

\bibitem{lan2025gateformer}
\BIBentryALTinterwordspacing
Y.-H. Lan, ``Gateformer: Advancing multivariate time series forecasting via temporal and variate-wise attention with gated representations,'' in \emph{1st ICML Workshop on Foundation Models for Structured Data}, 2025. [Online]. Available: \url{https://openreview.net/forum?id=wufwo0hkWr}
\BIBentrySTDinterwordspacing

\bibitem{li2018dcrnn_traffic}
Y.~Li, R.~Yu, C.~Shahabi, and Y.~Liu, ``Diffusion convolutional recurrent neural network: Data-driven traffic forecasting,'' in \emph{International Conference on Learning Representations (ICLR '18)}, 2018.

\bibitem{yu2018spatio}
B.~Yu, H.~Yin, and Z.~Zhu, ``Spatio-temporal graph convolutional networks: a deep learning framework for traffic forecasting,'' in \emph{Proceedings of the 27th International Joint Conference on Artificial Intelligence}, 2018, pp. 3634--3640.

\bibitem{wu2020connecting}
Z.~Wu, S.~Pan, G.~Long, J.~Jiang, X.~Chang, and C.~Zhang, ``Connecting the dots: Multivariate time series forecasting with graph neural networks,'' in \emph{Proceedings of the 26th ACM SIGKDD international conference on knowledge discovery \& data mining}, 2020, pp. 753--763.

\bibitem{zheng2020gman}
C.~Zheng, X.~Fan, C.~Wang, and J.~Qi, ``Gman: A graph multi-attention network for traffic prediction,'' in \emph{Proceedings of the AAAI conference on artificial intelligence}, vol.~34, no.~01, 2020, pp. 1234--1241.

\bibitem{guo2021learning}
S.~Guo, Y.~Lin, H.~Wan, X.~Li, and G.~Cong, ``Learning dynamics and heterogeneity of spatial-temporal graph data for traffic forecasting,'' \emph{IEEE Transactions on Knowledge and Data Engineering}, vol.~34, no.~11, pp. 5415--5428, 2021.

\bibitem{lyu2025autostf}
T.~Lyu, W.~Zhang, J.~Deng, and H.~Liu, ``Autostf: Decoupled neural architecture search for cost-effective automated spatio-temporal forecasting,'' in \emph{Proceedings of the 31st ACM SIGKDD Conference on Knowledge Discovery and Data Mining V. 1}, 2025, pp. 985--996.

\bibitem{kong2024spatio}
W.~Kong, Z.~Guo, and Y.~Liu, ``Spatio-temporal pivotal graph neural networks for traffic flow forecasting,'' in \emph{Proceedings of the AAAI conference on artificial intelligence}, vol.~38, no.~8, 2024, pp. 8627--8635.

\bibitem{bai2020adaptive}
L.~Bai, L.~Yao, C.~Li, X.~Wang, and C.~Wang, ``Adaptive graph convolutional recurrent network for traffic forecasting,'' \emph{Advances in neural information processing systems}, vol.~33, pp. 17\,804--17\,815, 2020.

\bibitem{shang2021discrete}
C.~Shang and J.~Chen, ``Discrete graph structure learning for forecasting multiple time series,'' in \emph{Proceedings of International Conference on Learning Representations}, 2021.

\bibitem{cao2020spectral}
D.~Cao, Y.~Wang, J.~Duan, C.~Zhang, X.~Zhu, C.~Huang, Y.~Tong, B.~Xu, J.~Bai, J.~Tong \emph{et~al.}, ``Spectral temporal graph neural network for multivariate time-series forecasting,'' \emph{Advances in neural information processing systems}, vol.~33, pp. 17\,766--17\,778, 2020.

\bibitem{li2024dynamic}
Y.~Li, Z.~Shao, Y.~Xu, Q.~Qiu, Z.~Cao, and F.~Wang, ``Dynamic frequency domain graph convolutional network for traffic forecasting,'' in \emph{ICASSP 2024-2024 IEEE International Conference on Acoustics, Speech and Signal Processing (ICASSP)}.\hskip 1em plus 0.5em minus 0.4em\relax IEEE, 2024, pp. 5245--5249.

\bibitem{zeng2023transformers}
A.~Zeng, M.~Chen, L.~Zhang, and Q.~Xu, ``Are transformers effective for time series forecasting?'' in \emph{Proceedings of the AAAI conference on artificial intelligence}, vol.~37, no.~9, 2023, pp. 11\,121--11\,128.

\bibitem{das2023longterm}
\BIBentryALTinterwordspacing
A.~Das, W.~Kong, A.~Leach, S.~K. Mathur, R.~Sen, and R.~Yu, ``Long-term forecasting with ti{DE}: Time-series dense encoder,'' \emph{Transactions on Machine Learning Research}, 2023. [Online]. Available: \url{https://openreview.net/forum?id=pCbC3aQB5W}
\BIBentrySTDinterwordspacing

\bibitem{zhou2021informer}
H.~Zhou, S.~Zhang, J.~Peng, S.~Zhang, J.~Li, H.~Xiong, and W.~Zhang, ``Informer: Beyond efficient transformer for long sequence time-series forecasting,'' in \emph{Proceedings of the AAAI conference on artificial intelligence}, vol.~35, no.~12, 2021, pp. 11\,106--11\,115.

\bibitem{wu2021autoformer}
H.~Wu, J.~Xu, J.~Wang, and M.~Long, ``Autoformer: Decomposition transformers with auto-correlation for long-term series forecasting,'' \emph{Advances in neural information processing systems}, vol.~34, pp. 22\,419--22\,430, 2021.

\bibitem{zhou2022fedformer}
T.~Zhou, Z.~Ma, Q.~Wen, X.~Wang, L.~Sun, and R.~Jin, ``Fedformer: Frequency enhanced decomposed transformer for long-term series forecasting,'' in \emph{International conference on machine learning}.\hskip 1em plus 0.5em minus 0.4em\relax PMLR, 2022, pp. 27\,268--27\,286.

\bibitem{zhang2023crossformer}
Y.~Zhang and J.~Yan, ``Crossformer: Transformer utilizing cross-dimension dependency for multivariate time series forecasting,'' in \emph{The eleventh international conference on learning representations}, 2023.

\bibitem{liu2024itransformer}
\BIBentryALTinterwordspacing
Y.~Liu, T.~Hu, H.~Zhang, H.~Wu, S.~Wang, L.~Ma, and M.~Long, ``itransformer: Inverted transformers are effective for time series forecasting,'' in \emph{The Twelfth International Conference on Learning Representations}, 2024. [Online]. Available: \url{https://openreview.net/forum?id=JePfAI8fah}
\BIBentrySTDinterwordspacing

\bibitem{li2024towards}
G.~Li, Z.~Zhao, X.~Guo, L.~Tang, H.~Zhang, and J.~Wang, ``Towards integrated and fine-grained traffic forecasting: A spatio-temporal heterogeneous graph transformer approach,'' \emph{Information Fusion}, vol. 102, p. 102063, 2024.

\bibitem{li2025fine}
S.~Li, W.~Yang, Y.~Cui, X.~Liu, L.~Meng, L.~Ma, and F.~Zhang, ``Fine-grained traffic inference from road to lane via spatio-temporal graph node generation,'' in \emph{Proceedings of the 31st ACM SIGKDD Conference on Knowledge Discovery and Data Mining V. 2}, 2025, pp. 1529--1540.

\bibitem{li2020fourier}
Z.~Li, N.~Kovachki, K.~Azizzadenesheli, B.~Liu, K.~Bhattacharya, A.~Stuart, and A.~Anandkumar, ``Fourier neural operator for parametric partial differential equations,'' \emph{arXiv preprint arXiv:2010.08895}, 2020.

\bibitem{nag2026spatio}
P.~Nag, A.~Zammit-Mangion, S.~Singh, and N.~Cressie, ``Spatio-temporal modeling and forecasting with fourier neural operators,'' \emph{arXiv preprint arXiv:2601.01813}, 2026.

\bibitem{zhang2025novel}
X.~Zhang, C.~Zhang, X.~Cai, Y.~Zhu, and C.~Luo, ``A novel spatiotemporal fourier neural operator for dynamic wake prediction,'' \emph{Energy}, p. 139233, 2025.

\bibitem{lin2026fano}
K.~Lin, H.~Lin, B.~Zhang, Y.~Ge, D.~Liu, X.~Li, Y.~Ye, and C.~Luo, ``Fano: Fourier advection neural operator for weather prediction,'' \emph{IEEE Transactions on Geoscience and Remote Sensing}, 2026.

\bibitem{yi2024filternet}
K.~Yi, J.~Fei, Q.~Zhang, H.~He, S.~Hao, D.~Lian, and W.~Fan, ``Filternet: Harnessing frequency filters for time series forecasting,'' \emph{Advances in Neural Information Processing Systems}, vol.~37, pp. 55\,115--55\,140, 2024.

\bibitem{eldele2024tslanet}
E.~Eldele, M.~Ragab, Z.~Chen, M.~Wu, and X.~Li, ``Tslanet: rethinking transformers for time series representation learning,'' in \emph{Proceedings of the 41st International Conference on Machine Learning}, 2024, pp. 12\,409--12\,428.

\bibitem{yi2023fouriergnn}
K.~Yi, Q.~Zhang, W.~Fan, H.~He, L.~Hu, P.~Wang, N.~An, L.~Cao, and Z.~Niu, ``Fouriergnn: Rethinking multivariate time series forecasting from a pure graph perspective,'' \emph{Advances in neural information processing systems}, vol.~36, pp. 69\,638--69\,660, 2023.

\bibitem{koren2003spectral}
Y.~Koren, ``On spectral graph drawing,'' in \emph{International Computing and Combinatorics Conference}.\hskip 1em plus 0.5em minus 0.4em\relax Springer, 2003, pp. 496--508.

\bibitem{sun2025human}
P.~Sun, J.~Yang, Z.~Huang, S.~Huang, S.~Xu, W.~Wang, W.~Guo, Y.~Feng, X.~Zhai, T.~Yang \emph{et~al.}, ``Human mobility datasets in the complex metro system of shanghai,'' \emph{Scientific Data}, vol.~12, no.~1, p. 1061, 2025.

\bibitem{shao2024exploring}
Z.~Shao, F.~Wang, Y.~Xu, W.~Wei, C.~Yu, Z.~Zhang, D.~Yao, T.~Sun, G.~Jin, X.~Cao \emph{et~al.}, ``Exploring progress in multivariate time series forecasting: Comprehensive benchmarking and heterogeneity analysis,'' \emph{IEEE Transactions on Knowledge and Data Engineering}, vol.~37, no.~1, pp. 291--305, 2024.

\bibitem{gao2024spatial}
H.~Gao, R.~Jiang, Z.~Dong, J.~Deng, Y.~Ma, and X.~Song, ``Spatial-temporal-decoupled masked pre-training for spatiotemporal forecasting,'' in \emph{Proceedings of the Thirty-Third International Joint Conference on Artificial Intelligence}, 2024.

\bibitem{nie2023a}
\BIBentryALTinterwordspacing
Y.~Nie, N.~H. Nguyen, P.~Sinthong, and J.~Kalagnanam, ``A time series is worth 64 words: Long-term forecasting with transformers,'' in \emph{The Eleventh International Conference on Learning Representations}, 2023. [Online]. Available: \url{https://openreview.net/forum?id=Jbdc0vTOcol}
\BIBentrySTDinterwordspacing

\bibitem{li2013efficient}
L.~Li, Y.~Li, and Z.~Li, ``Efficient missing data imputing for traffic flow by considering temporal and spatial dependence,'' \emph{Transportation research part C: emerging technologies}, vol.~34, pp. 108--120, 2013.

\end{thebibliography}

% \newpage

 

% \end{IEEEbiography}



\vfill

\end{document}